\documentclass[runningheads]{llncs}
\usepackage[dvipsnames,table,xcdraw]{xcolor}
\usepackage[T1]{fontenc}
\usepackage{graphicx}

\usepackage{amssymb} 
\usepackage{amsmath}
\usepackage{multirow}
\usepackage{subcaption}
\usepackage{enumitem}
\usepackage[deletedmarkup=sout, color=red]{changes}
\definecolor{BdazzledBlue}{HTML}{2e58a5}
\newcommand{\indicator}[1]{\mathbf{1}}

\definecolor{cvprblue}{rgb}{0.21,0.49,0.74}

\usepackage[pagebackref,breaklinks,colorlinks,citecolor=cvprblue]{hyperref}

\DeclareMathOperator*{\argmin}{arg\,min}

\begin{document}
\title{CloudFort: Enhancing Robustness of 3D Point Cloud Classification Against Backdoor Attacks via Spatial Partitioning and Ensemble Prediction}
\titlerunning{CloudFort}
%
\author{Wenhao Lan\inst{1} \and
Yijun Yang\inst{2} \and Haihua Shen\inst{1} \thanks{Corresponding author.} \and Shan Li\inst{1} }

\authorrunning{W. Lan et al.}
%
\institute{University of Chinese Academy of Sciences, Beijing 100049, China\\ 
\email{\{lanwenhao21,lishan203\}@mails.ucas.ac.cn}\\
\email{shenhh@ucas.ac.cn}
\and
The Chinese University of Hong Kong, Shatin, N.T., Hong Kong 
\email{yjyang@cse.cuhk.edu.hk}}
\maketitle              
\begin{abstract}
The increasing adoption of 3D point cloud data in various applications, such as autonomous vehicles, robotics, and virtual reality, has brought about significant advancements in object recognition and scene understanding. However, this progress is accompanied by new security challenges, particularly in the form of backdoor attacks. These attacks involve inserting malicious information into the training data of machine learning models, potentially compromising the model's behavior. In this paper, we propose CloudFort, a novel defense mechanism designed to enhance the robustness of 3D point cloud classifiers against backdoor attacks. CloudFort leverages spatial partitioning and ensemble prediction techniques to effectively mitigate the impact of backdoor triggers while preserving the model's performance on clean data. We evaluate the effectiveness of CloudFort through extensive experiments, demonstrating its strong resilience against the Point Cloud Backdoor Attack (PCBA). Our results show that CloudFort significantly enhances the security of 3D point cloud classification models without compromising their accuracy on benign samples. Furthermore, we explore the limitations of CloudFort and discuss potential avenues for future research in the field of 3D point cloud security. The proposed defense mechanism represents a significant step towards ensuring the trustworthiness and reliability of point-cloud-based systems in real-world applications.

\end{abstract}
\section{Introduction}
\label{sec:intro}
The rapid growth of 3D point cloud data has revolutionized various fields, including autonomous vehicles~\cite{maturana2018real,rajathi2023path,xie2023real,zheng2022global,chen20203d,cui2021deep}, robotics~\cite{chen2022direct,duan2021robotics}, and virtual reality~\cite{blanc2020genuage,alexiou2020pointxr}. The increasing adoption of 3D point clouds in these applications has brought about significant advancements in object recognition and scene understanding, with point cloud classification being a fundamental task. Several deep learning-based models have been proposed to tackle this task, which aims to assign semantic labels to 3D point clouds~\cite{qi2017pointnet,qi2017pointnet++,wang2019dynamic}. However, this progress is accompanied by new security challenges.

One significant threat to the integrity of these models is backdoor attacks \cite{xiang2021backdoor,li2021pointba,gao2023imperceptible,wen2021generative,wu2023computation}. In a backdoor attack, an adversary manipulates the training data by inserting malicious samples that are designed to trigger a specific behavior in the model under certain conditions. The Point Cloud Backdoor Attack (PCBA) \cite{xiang2021backdoor}, as a representative backdoor attack for 3D point-cloud models injects a cluster of malicious points into the training data, which serve as the backdoor triggers.
When the trigger is present in the input data during inference, the model misclassifies the point cloud to a target class chosen by the attacker. The stealthy nature of PCBA makes it particularly challenging to detect and defend against.

Current research on defense mechanisms against backdoor attacks primarily focuses on the image domain~\cite{doan2020februus,wu2021adversarial,dong2021black,xu2021detecting,borgnia2021strong,zeng2021rethinking,weber2023rab}, and these methods cannot be directly applied to 3D point cloud data due to the inherent differences in data representation and structure. Recently, some researchers have proposed defense methods specifically designed for 3D point cloud classifiers\cite{xiang2022detecting,li2024pointcvar,hu2023pointcrt}. However, these methods have limitations. For instance, Xiang et al.\cite{xiang2022detecting} proposed a reverse-engineering method to detect backdoors in point cloud models, but it requires significant computational resources and time to identify and remove the poisoned samples from the training data and retrain the classifier. PointCVaR\cite{li2024pointcvar}, while effective in filtering out various types of point cloud outliers like random noise and adversarial noise, has limited defense efficacy against backdoor triggers. PointCRT\cite{hu2023pointcrt} can detect backdoor samples during the inference stage without prior knowledge of the trigger patterns, but it relies on the availability of clean samples as a reference and cannot be applied in scenarios where only a poisoned dataset is accessible.

\begin{figure}[tpb]
\centering
\includegraphics[width=\textwidth]{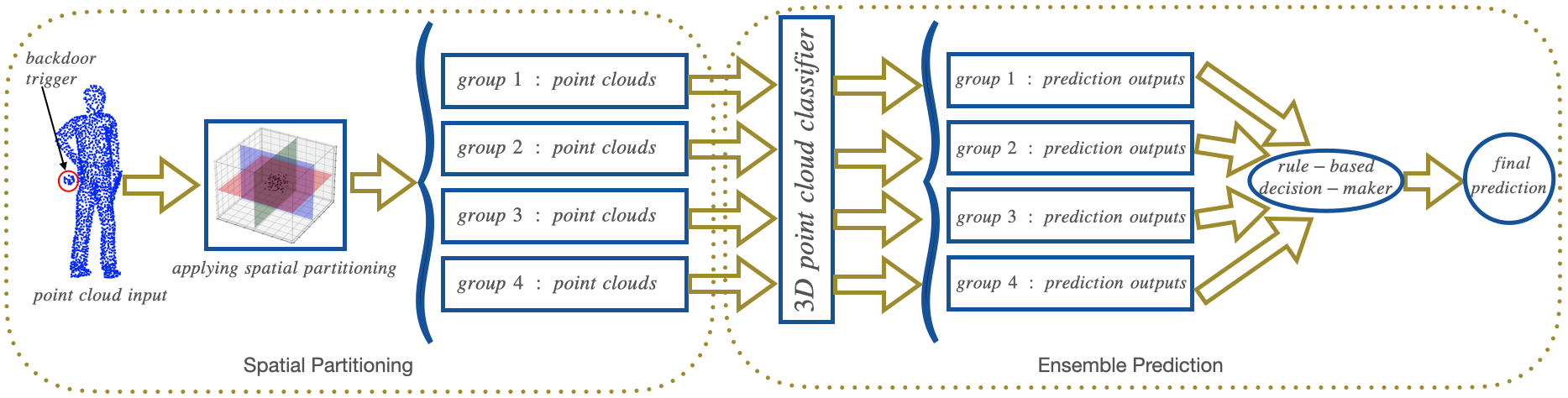}
\caption{Overview of the proposed CloudFort defense mechanism. }
\label{fig:cloudfort}
\end{figure}

To address the aforementioned limitations, we introduce CloudFort, a novel defense framework specifically designed to mitigate backdoor attacks in 3D point-cloud models. Unlike existing approaches that either incur significant computational costs~\cite{xiang2022detecting} or exhibit limited effectiveness against backdoor triggers~\cite{li2024pointcvar}, CloudFort stands out as a lightweight yet highly effective solution. CloudFort capitalizes on the inherent geometric properties of point clouds and harnesses the strength of ensemble learning techniques. This unique combination enables the framework to defend against backdoors from both spatial and multi-view prediction perspectives. By leveraging these key features, CloudFort provides robust defense while maintaining efficiency, making it a promising solution in the fight against backdoor attacks in 3D point-cloud models. More specifically, CloudFort employs two key defensive paradigms: trigger-backdoor mismatch and trigger elimination. As demonstrated in Figure~\ref{fig:cloudfort}, CloudFort employs a two-stage defense strategy: spatial partitioning and ensemble prediction. In the spatial partitioning stage, CloudFort first divides the 3D space that encompasses the input point cloud into eight equal regions and then generates eight sub-point clouds by sequentially removing points from each octant. some of which may not contain the backdoor trigger while still retaining the essential features of the original point cloud. To further enhance the likelihood of eliminating backdoor patterns, CloudFort employs four spatial partitioning strategies, resulting in four groups of sub-point clouds. The ensemble prediction technique is then applied to these sub-point clouds, leveraging multiple independent information sources to improve the accuracy and reliability of the final prediction.

In summary, the main contributions of this paper are as follows:
\begin{itemize}[leftmargin=*, itemsep=3pt]
\item We propose CloudFort, a direct defense framework designed to enhance the robustness of 3D point cloud classification models against backdoor attacks.
\item We introduce a novel and effective two-stage defense strategy in CloudFort, consisting of spatial partitioning and ensemble prediction.
\item We conduct extensive experiments to evaluate the effectiveness of CloudFort against PCBA, demonstrating its strong robustness and ability to preserve the model's performance on clean data.
\end{itemize}

\section{Related Work}
\label{sec:related work}

\subsection{3D Point Cloud and Classification Models}

A 3D point cloud comprises a collection of points, denoted as $P=\{p_i|\ p_i\in \mathbb{R}^3, i=1,...,n\}$, which represents the surface of a three-dimensional object. Each point $p_i$ is characterized by its 3D coordinates $(x_i,y_i,z_i)$. Point clouds are acquired through various sensing devices, including LiDAR scanners~\cite{di2021mobile,raj2020survey} and depth cameras~\cite{wu2020depth,omarali2020virtual}. Compared to 2D images, 3D point clouds offer several advantages. First, point clouds provide explicit 3D geometric information, allowing for more accurate and detailed representation of object shapes and scene structures. Second, point clouds are more robust to lighting variations and can capture data in low-light or even dark conditions, making them suitable for a wider range of scenarios. Third, point clouds enable direct 3D measurements and analyses, such as distance estimation and object segmentation, which are crucial for tasks like autonomous navigation and robotic manipulation. Furthermore, the sparse and irregular nature of point clouds offers the potential for efficient processing and storage compared to dense 2D pixel grids. Owing to these advantages, 3D point clouds have been widely adopted in various applications, including autonomous driving~\cite{maturana2018real,rajathi2023path,xie2023real,zheng2022global,chen20203d,cui2021deep}, robotics~\cite{chen2022direct,duan2021robotics}, and virtual reality~\cite{blanc2020genuage,alexiou2020pointxr}.

Several models have been proposed to tackle the task of point cloud classification. One notable model in this field is PointNet~\cite{qi2017pointnet}, which directly processes raw point clouds and extracts global features using point-wise MLPs and max-pooling. Building on PointNet, PointNet++\cite{qi2017pointnet++} introduces a hierarchical structure that captures local features at different scales, further improving performance. Another noteworthy model, DGCNN\cite{wang2019dynamic}, treats point clouds as graphs and employs graph convolution to learn both local and global features. These models have demonstrated state-of-the-art performance and have gained significant popularity in practical applications.

\subsection{Backdoor Attack}
Backdoor attack aim to manipulate the model's behavior by poisoning the training data with a trigger pattern, causing the model to misclassify triggered samples into a target class~\cite{li2022backdoor}. Inspired by the success of backdoor attacks in the image domain~\cite{gu2017badnets,nguyen2021wanet,liu2020reflection,turner2019label}, researchers have started to explore backdoor attacks on point cloud data.

The Point Cloud Backdoor Attack (PCBA)~\cite{xiang2021backdoor} introduce backdoors in 3D point cloud classifiers. Experiments show that PCBA achieves high attack success rates($ \geq 87\%$
) on PointNet, PointNet++ and DGCNN, indicating that the attacker can arbitrarily manipulate the victim model's predictions when point clouds with the backdoor trigger are present. Such high attack success rates pose serious threats to the reliability and robustness of 3D point cloud models in safety-critical applications such as autonomous driving, where an attacker could potentially induce the model to misidentify pedestrians as non-pedestrians and cause severe safety incidents. The basic idea of PCBA is injecting a cluster of points as the trigger into the point cloud, aiming to mislead the classifier to predict a target class. Formally, let $\mathcal{B}(P;V)$ denote the backdoor trigger injection function, where $P=\{p_i|\ p_i\in \mathbb{R}^3, i=1,...,n\}$ is the original point cloud and $V=\{u_i+c|\ u_i\in \mathbb{R}^3, c\in \mathbb{R}^3, i=1,...,n'\}$ is the trigger. The local geometry of the trigger $u_i$ and its spatial location $c$ are jointly optimized to ensure a high attack success rate and stealthiness. Let $\mathcal{F}_v(\cdot;\Theta):\mathcal{P}\rightarrow \mathcal{Y}$ denote the victim point cloud classifier with weights $\Theta$, which predicts the label of a given point cloud. PCBA optimize the objective function Eq.~\ref{eq:objective} to alert the prediction from source class $s$ to target class $t$:
\begin{equation}
\label{eq:objective}
E_{P\sim H_s}[\indicator{}(\mathcal{F}_v(\mathcal{B}(P;V);\Theta)=t)]
\end{equation}
where $H_s$ is the distribution of point clouds from source class $s$, $t$ is the attacker's target class, and $\mathbf{1}(\cdot)$ is the indicator function. 

Several subsequent works have further explored the design space of backdoor attacks on point clouds. PointBA~\cite{li2021pointba} introduced a unified framework with spatial transform and orientation-based triggers, but overlooked the intrinsic rotational invariance in such classifier~\cite{qi2017pointnet++}, limiting its practicality. IRBA~\cite{gao2023imperceptible} employs weighted local transformations to generate unique non-linear deformations that can bypass preprocessing, but relies on fixed hyper-parameters, potentially limiting its applicability. Wu et al.~\cite{wu2023computation} propose a computation and data-efficient method by selecting the most critical poisoned samples, while other exploratory efforts exist~\cite{zhang2022towards,wen2021generative}.

\subsection{Backdoor Defenses}

Existing defense frameworks predominantly target backdoor attacks on image data, employing a variety of strategies. These include denoising input images via diffusion models~\cite{nie2022diffusion,lee2023robust}, removing triggers based on their distinctive patterns~\cite{doan2020februus,qiu2021deepsweep}, and fine-tuning models to counteract anomalous behaviors induced by backdoors~\cite{yoshida2020disabling,wu2021adversarial}. While these methods demonstrate efficacy in defending against backdoor attacks within the image domain, they exhibit limited generalizability to the 3D point cloud domain~\cite{xiang2021backdoor,li2021pointba,gao2023imperceptible}. Moreover, they may significantly impair the original performance of the systems. Consequently, the defense against backdoor attacks in the 3D point cloud field remains a largely under-explored area.

To address the issue of backdoor attacks in 3D point clouds, Xiang et al.~\cite{xiang2022detecting} proposed a reverse-engineering approach to detect backdoors in point cloud models. Their method optimizes the loss function of the target model and then estimates the backdoor pattern for each source class. Moreover, they introduced a composite detection statistic to reduce false positives. However, a significant limitation of their defense is that even when a backdoored model is detected, it cannot immediately provide correct predictions. Instead, it necessitates identifying and removing the poisoned samples from the training dataset and retraining the point cloud classifier, which demands substantial computational resources and incurs significant time overhead. Li et al.~\cite{li2024pointcvar} introduced an innovative point cloud outlier removal method called PointCVar, which leverages downstream classification models to define the concept of point risk through gradient-based attribution analysis and reformulates the outlier removal process as an optimization problem. Although this method demonstrates excellent performance in filtering out various types of point cloud outliers, it primarily focuses on random noise and adversarial noise, exhibiting limited effectiveness in defending against backdoor triggers. Hu et al. ~\cite{hu2023pointcrt} proposed the PointCRT method, which utilizes a model's robustness to different corruptions to detect backdoor samples during the inference stage, without requiring prior knowledge of the backdoor trigger patterns. While this black-box detection method can effectively handle multiple types of 3D backdoor attacks, its main limitation lies in the need for clean samples as references, rendering it inapplicable in scenarios where only a poisoned dataset is available.

Another research line in this domain is the PointGuard framework proposed by \cite{liu2021pointguard}. It employs random sub-sampling and majority voting to achieve provable robustness against various types of adversarial attacks on point clouds. While the defense performance is commendable, it is worth noting that it often leads to an unacceptable performance drop on normal 3D point-cloud input data in many real-world applications.

To overcome these aforementioned limitations, CloudFort, our proposed defense mechanism, leverages the power of spatial partitioning and ensemble prediction to enhance the robustness of 3D point cloud classifiers against backdoor attacks. By employing spatial partitioning, CloudFort effectively mitigates the influence caused by triggers, while the ensemble prediction strategy ensures that the overall performance remains intact, without compromising the standard performance. This approach not only provides an effective defense mechanism but also ensures efficiency, making it suitable for real-world deployment scenarios.

\section{Method}
\label{sec:method}

\subsection{Overview}
Contrasting with those conventional methods, this study introduces an innovative defense strategy CloudFort that can deduce accurate predictive outcomes even in scenarios where the model has been compromised by a backdoor. This capability not only enhances the resilience of point cloud processing systems against PCBA\cite{xiang2021backdoor} but also optimizes resource utilization by circumventing the need for data re-collection and model re-training.

\begin{figure}[tpb]
\centering
\includegraphics[width=\textwidth]{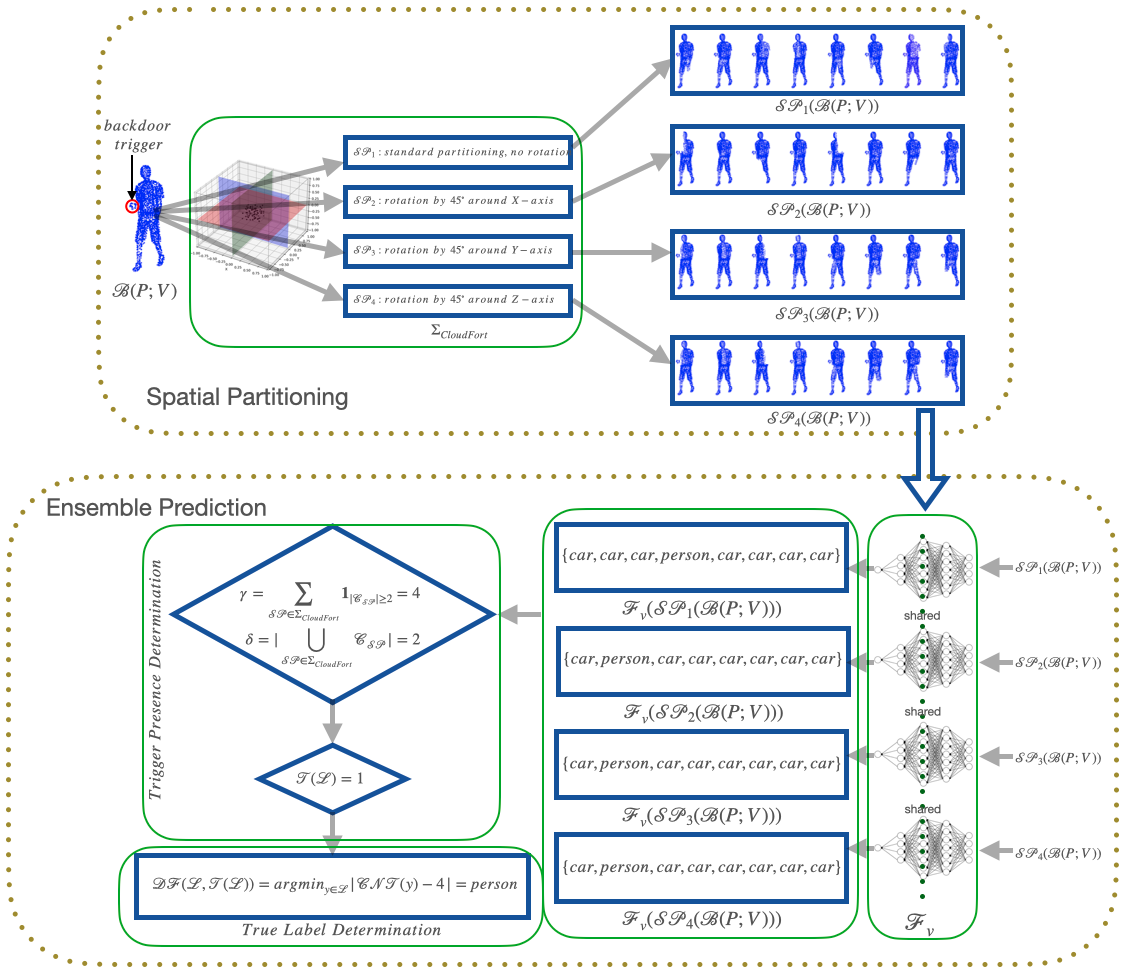}
\caption{Schematic representation of the CloudFort defense mechanism detailing the sequential Process of Spatial Partitioning (Sec. \ref{subsec:spatial partitioning}) and Ensemble Prediction (Sec. \ref{subsec:ensemble prediction}) for robust backdoor attack mitigation in 3D point cloud classification, illustrated with an example attack scenario where the source class is 'person' and the target class is 'car'. The input point cloud, potentially containing a backdoor trigger, undergoes spatial partitioning ($\Sigma_{CloudFort}$), generating multiple sub-point clouds by systematically excluding different regions. These sub-point clouds are then processed by the compromised classifier ($\mathcal{F}_v$), yielding a set of predictions. The trigger presence determination module ($\mathcal{T}$) analyzes the prediction patterns to infer the existence of a backdoor. Finally, the true label determination module ($\mathcal{DF}$) leverages the trigger presence information and the prediction statistics to deduce the correct classification label ($y_{true}$), effectively mitigating the impact of the backdoor attack and correctly classifying the input as 'person' instead of the attacker's target class 'car'.}
\label{fig:defense_against_PCBA}
\end{figure}

To achieve this objective, CloudFort, integrates the synergistic benefits of both trigger-backdoor mismatch and trigger elimination defense paradigms. This strategy is meticulously formulated, relying on the combined application of spatial partitioning and ensemble prediction techniques. This dual-pronged approach synergistically constitutes a robust framework, purposefully architected to mitigate the threats posed by such attacks effectively. By leveraging the spatial partitioning technique, CloudFort systematically segments the 3D space into discrete, manageable units, thereby isolating and neutralizing potential backdoor triggers embedded within the point cloud data. Concurrently, the ensemble prediction mechanism harnesses the collective predictive capabilities of multiple models, enhancing the accuracy and reliability of outcomes even in the presence of compromised data. Together, these paradigms form the cornerstone of CloudFort's methodology, offering a comprehensive solution to fortify point cloud processing systems against the vulnerabilities exploited by backdoor attacks, as shown in Figure \ref{fig:defense_against_PCBA}.

\subsection{Objective, assumption and key intuition}
\subsubsection{Objective.} The primary aim of this research is to ensure that the defensive mechanism, CloudFort, maintains the integrity of predictive outcomes in scenarios where the prediction model has been compromised by the insertion of a backdoor. Formally, let \(\mathcal{F}_v(\mathcal{B}(P;V);\Theta)\) represent a prediction model that has been subjected to a backdoor attack, and let \( \mathcal{S} \) denote a set of input samples, where each sample \( P_i \in \mathcal{S} \) may or may not carry a backdoor trigger \( V \). The objective can be expressed as follows: Consider CloudFort as a novel prediction function, denoted \( \mathcal{F}_{CloudFort} \) . for every input sample \( P_i \), regardless of the presence of \( V \), CloudFort aims to accurately predict the correct classification label \( y_{true} \) ,  such that the prediction function \( \mathcal{F}_{CloudFort}(\mathcal{F}_v(\mathcal{B}(P_i;V);\Theta)) = y_{true} \), ensuring that \( \mathcal{F}_{CloudFort} \) exhibits resilience against the adversarial manipulation introduced by \( V\). This ensures that CloudFort effectively neutralizes the influence of any backdoor triggers, thereby preserving the fidelity of the model's predictions across all input samples within \( \mathcal{S} \).

\subsubsection{Assumption.} In this study, several foundational assumptions are posited regarding the defender's capabilities and resources. Firstly, it is assumed that the defender does not have access to the classifier's training dataset, precluding the possibility of analyzing or leveraging this dataset for defensive purposes. Secondly, the defender is presumed to lack the capability to train any classifier, which implies an inability to employ traditional retraining or model refinement strategies as a means of counteracting adversarial threats. Lastly, the absence of a clean dataset is stipulated, indicating that the defender cannot utilize untainted data for the purpose of model validation or the establishment of a baseline for detecting anomalous model behavior. These assumptions collectively delineate the constraints within which the defense strategy must operate, underscoring the necessity for innovative approaches that do not rely on direct interaction with the training process or the availability of pristine data.

\subsubsection{Intuition.} In the context of this research, the defense strategy against backdoor attacks in point cloud classification is predicated on several intuitive premises. The successful implementation of CloudFort substantiates the validity of these intuitions. \textbf{I1:} Feature Retention in 2.5D Scenarios. The term "2.5D," within the context of point cloud analysis, refers to scenarios characterized by partial occlusion. Consider a point cloud \( P \) subjected to a spatial partitioning function \( \mathcal{SP}(P) \) which simulates partial occlusion by segmenting and removing a portion of the point cloud, It is hypothesized that:
\begin{equation}
\forall P, \exists \mathcal{FT}_{\text{essential}}(P) \subseteq \mathcal{FT}_{\text{essential}}(\mathcal{SP}(P))
\end{equation}
where \( \mathcal{FT}_{\text{essential}}(P) \) denotes the set of essential features within \( P \) necessary for accurate classification. This implies that the classifier \( \mathcal{F} \) can extract necessary classification features from \( \mathcal{SP}(P) \). \textbf{I2:}Trigger Mitigation through Spatial Partitioning. By applying a spatial partitioning function \( \mathcal{SP}(P) \) that segments and removes a portion of the point cloud, there exists a probability \( \Pr \) that the embedded trigger \( V \) within \( P \) is removed, thereby potentially neutralizing the backdoor effect. Formally:
\begin{equation}
     \Pr(V \notin \mathcal{SP}(P)) > 0
\end{equation}
suggesting that spatial partitioning may inadvertently excise the portion containing \( V \). \textbf{I3:} Diverse Partitioning Strategies and Prediction Analysis. Employing various spatial partitioning strategies \( \{\mathcal{SP}_1, \mathcal{SP}_2, \ldots, \mathcal{SP}_n\} \) generates a series of 2.5D point clouds \( \{P_1, P_2, \ldots, P_m\} \). When these modified point clouds are processed through a classifier \( \mathcal{F} \) compromised with a backdoor, the resulting set of prediction labels \( \{y_1^*, y_2^*, \ldots, y_m^*\} \) may exhibit distinguishable patterns that:
\begin{equation}
\mathcal{DF}(\{y_1^*, y_2^*, \ldots, y_m^*\}) \rightarrow y_{\text{true}}
\end{equation}
where \( \mathcal{DF} \) denotes a decision function capable of deducing the correct classification \( y_{\text{true}} \) from the observed label patterns. This premise suggests that the analysis of prediction variability across differently partitioned samples could facilitate the identification of the correct classification outcome, notwithstanding the backdoor manipulation. 

\subsubsection{Defense procedure.}The defensive mechanism of CloudFort is orchestrated through a sequential integration of spatial partitioning followed by ensemble prediction.

\subsection{Step 1: Spatial Partitioning}
\label{subsec:spatial partitioning}

Informed by Intuition 1 and 2—namely, that a 2.5D representation of a point cloud retains critical feature information for classification, and that strategic spatial partitioning may excise embedded triggers—this study needs an optimized spatial partitioning strategy as a key component of the CloudFort defense mechanism. 

Let \( P\) represent the original point cloud in a 3D space \( \mathbb{R}^3 \). The spatial partitioning function, \( \mathcal{SP} \), is defined to segment \( P \) into \( n \) partitions:
\begin{equation}
\mathcal{SP}(P) = \{P_{\text{sub}_1}, P_{\text{sub}_2}, \ldots, P_{\text{sub}_n}\}
\end{equation}
Each partition \( P_i \) represents the point cloud subset after the removal of the \( i \)-th segment, which could potentially contain the trigger.

Inspired by the application of Octree in the processing of 3D point clouds\cite{vo2015octree}, the spatial partitioning function, \( \mathcal{SP} \),  is achieved through the implementation of cutting planes aligned with the coordinate system's axes, resulting in the segmentation of space into eight unique regions, denoted as \( \{R_1, R_2, \ldots, R_8\} \).

For each iteration of the partitioning process, one specific region \( R_i \) is systematically excluded, and the point clouds contained within the remaining seven regions are aggregated to form a sub-point cloud \( P_{\text{sub}_i} \). Formally, this process can be represented as:
\begin{equation}
P_{\text{sub}_i} = \bigcup_{\substack{j=1 \\ j \neq i}}^{8} P \cap R_j, \quad \text{for } i = 1, 2, \ldots, 8
\end{equation}

This operation yields a collection of eight sub-point clouds: \(P_{\text{sub}_1}\), \(P_{\text{sub}_2}\), \ldots, \(P_{\text{sub}_8}\). Each derived by excluding the point cloud data from one of the eight regions in turn. This strategic exclusion and aggregation process facilitates a comprehensive analysis of the point cloud data from multiple perspectives, enhancing the robustness of subsequent predictive analyses. However, this operation inherently carries potential challenges that necessitate careful consideration.

\subsubsection{Challenge 1: Potential Removal of Critical Features.} When spatially partitioning a 3D point cloud, a nuanced consideration must be given to the potential exclusion of key point clusters that embody essential characteristics of the point cloud. These clusters, denoted as \( \mathcal{P}_{\text{critical}} \), are aggregations of points within \( P \) that encapsulate salient features indispensable for precise and accurate classification.

Formally, let \( \mathcal{P}_{\text{critical}} = \{P_{c_1}, P_{c_2}, \ldots, P_{c_k}\} \) be the set of key point clusters within \( P \), where each \( P_{c_j} \) represents a cluster of points that are collectively critical for maintaining the integrity of the classification outcome. The act of partitioning may result in scenarios where one or more of these critical clusters are entirely excluded from a partition \( P_{\text{sub}_i} \), effectively denoted as \( P_{c_j} \cap P_{\text{sub}_i} = \emptyset \) for any \( P_{c_j} \in \mathcal{P}_{\text{critical}} \) and some \( P_{\text{sub}_i} \). This exclusion poses a significant risk, as the absence of \( P_{c_j} \) from \( P_{\text{sub}_i} \) could lead to a misrepresentation of the original point cloud's intrinsic properties, thereby impairing the classification's fidelity and accuracy.

\subsubsection{Challenge 2: Introduction of Inherent Backdoors.} Moreover, the spatial partitioning strategy might inadvertently introduce new vulnerabilities, conceptualized as inherent backdoors. This paradox arises when the partitioning process \( \mathcal{SP} \) creates configurations \( V_{\mathcal{SP}} \) within the point cloud that were not present initially, potentially aligning with adversarial patterns recognized by the compromised classifier \( \mathcal{F} \). Such configurations might not trigger the original backdoor but could lead to misclassifications due to the altered structural integrity of \( P \), effectively acting as an unintended backdoor.

\subsubsection{Strategic Mitigation.} To mitigate these concerns, CloudFort adopts a multi-faceted spatial partitioning approach, employing a variety of strategies \( \Sigma = \{\mathcal{SP}_1, \mathcal{SP}_2, \ldots , \mathcal{SP}_k\} \) that introduce rotational dynamics to the partitioning planes. This diversity in partitioning methodologies aims to reduce the likelihood that all partitioning strategies \( \Sigma \) simultaneously remove critical features and introduce new vulnerabilities. 

Given a probability \( Pr(0 < Pr < 1 ) \) that a single partitioning operation might either remove critical features or introduce new vulnerabilities, the likelihood that all partitioning strategies \( \Sigma \) simultaneously succumb to these pitfalls is exponentially reduced, mathematically represented as \( pr^k \), where \( k \) is the number of strategies employed. This probabilistic safeguard ensures that there is a significantly reduced chance of all partitioning operations adversely affecting the model's integrity, thereby enhancing the robustness of CloudFort against such occurrences.

However, it's crucial to strike a balance between the diversity of partitioning strategies and computational efficiency. An excessive number of partitioning strategies can lead to increased computational complexity, adversely affecting the system's performance. To address this, CloudFort employs a set of spatial partitioning strategies \( \Sigma_{CloudFort} = \{\mathcal{SP}_1, \mathcal{SP}_2, \mathcal{SP}_3 , \mathcal{SP}_4\} \) :

\begin{enumerate}
    \item \( \mathcal{SP}_1 \) corresponds to the standard partitioning with no rotation.
    \item \( \mathcal{SP}_2 \) applies a rotation \( \mathcal{R}_{X}(45^\circ) \) about the X-axis to the partitioning planes from \( \mathcal{SP}_1 \).
    \item \( \mathcal{SP}_3 \) applies a rotation \( \mathcal{R}_{Y}(45^\circ) \) about the Y-axis to the partitioning planes from \( \mathcal{SP}_1 \).
    \item \( \mathcal{SP}_4 \) applies a rotation \( \mathcal{R}_{Z}(45^\circ) \) about the Z-axis to the partitioning planes from \( \mathcal{SP}_1 \).
\end{enumerate}

This set of strategies yield four groups of sub-point clouds. For a given \( \mathcal{SP} \in \Sigma_{CloudFort} \), the resultant sub-point clouds can be denoted as \( P_{\mathcal{SP}, i} \), where \( i \) ranges from 1 to 8, corresponding to each segment's systematic exclusion.

\subsection{Step 2: Ensemble Prediction}
\label{subsec:ensemble prediction}

Through Step 1 of the CloudFort methodology, by employing a set of spatial partitioning strategies \( \Sigma_{CloudFort}\), the process systematically generates a series of point cloud subsets. This operation yields four distinct groups of point clouds, each containing eight sub-point clouds, formalized as follows:
\begin{equation}
\Sigma_{CloudFort}(P) \rightarrow \left\{ \begin{matrix}
P_{\mathcal{SP}_1,1} & P_{\mathcal{SP}_1,2} & \cdots & P_{\mathcal{SP}_1,8} \\
P_{\mathcal{SP}_2,1} & P_{\mathcal{SP}_2,2} & \cdots & P_{\mathcal{SP}_2,8} \\
P_{\mathcal{SP}_3,1} & P_{\mathcal{SP}_3,2} & \cdots & P_{\mathcal{SP}_3,8} \\
P_{\mathcal{SP}_4,1} & P_{\mathcal{SP}_4,2} & \cdots & P_{\mathcal{SP}_4,8} \\
\end{matrix} \right\}    
\end{equation}

Based on intuition 3, With the generation of diverse 2.5D representations of the original point cloud \(P\) through the application of various spatial partitioning strategies $\Sigma_{CloudFort}$, the next phase involves consolidating the predictions from the compromised classifier $\mathcal{F}_\mathcal{V}$ on these modified inputs to deduce the correct classification label $y_{true}$. 

This consolidation is achieved through an ensemble methodology that analyzes the prediction patterns across the diverse inputs to nullify the influence of any triggers potentially present in \(P\). Concretely, for each sub-point cloud \(P_{\mathcal{SP}_j,i}\ (\forall\ \mathcal{SP}_j\in\Sigma_{CloudFort},\;i\in\{1, 2, \ldots, 8\})\), the prediction label $y^*_{\mathcal{SP}_j,i}$ is obtained using $\mathcal{F}_\mathcal{V}$:
\begin{equation}
\mathcal{F}_\mathcal{V}(\Sigma_{CloudFort}(P)) \rightarrow \left\{ \begin{matrix}
y^*_{\mathcal{SP}_1,1} & y^*_{\mathcal{SP}_1,2} & \cdots & y^*_{\mathcal{SP}_1,8} \\
y^*_{\mathcal{SP}_2,1} & y^*_{\mathcal{SP}_2,2} & \cdots & y^*_{\mathcal{SP}_2,8} \\
y^*_{\mathcal{SP}_3,1} & y^*_{\mathcal{SP}_3,2} & \cdots & y^*_{\mathcal{SP}_3,8} \\
y^*_{\mathcal{SP}_4,1} & y^*_{\mathcal{SP}_4,2} & \cdots & y^*_{\mathcal{SP}_4,8} \\
\end{matrix} \right\}
\end{equation}

The aggregation and analysis of these classification outcomes leverage the distinctive characteristics of the predictions to deduce the most accurate classification label for the original point cloud \( P \). This involves a systematic evaluation of the predictive consistency and reliability across the different spatially partitioned datasets, enabling CloudFort to identify the correct classification result amidst potential backdoor-induced misclassifications.

These labels are processed by a decision function $\mathcal{DF}$ designed to extract the correct classification from the observed prediction patterns, formalized as:
\begin{equation}
\mathcal{DF}(\mathcal{F}_\mathcal{V}(\Sigma_{CloudFort}(P))) \rightarrow y_{true}
\end{equation}

In this study, the decision function \( \mathcal{DF} \) is meticulously designed as a rule-based ensemble methodology, anchored on two pivotal principles that address the behavior of prediction outcomes based on the presence or absence of a backdoor trigger in the original point cloud \( P \):

\subsubsection{Consistency in Prediction for Non-Triggered Point Clouds.} When the original point cloud \( P \) does not contain a trigger, the predictions from the sub-point clouds within each group are expected to demonstrate a high degree of consistency. Mathematically, if \( P \) is trigger-free, then for any given partitioning strategy \( \mathcal{SP} \in \Sigma_{CloudFort} \), the resulting predictions \( \{y^*_{\mathcal{SP},1}, y^*_{\mathcal{SP},2}, \ldots, y^*_{\mathcal{SP},8}\} \) are likely to converge to a single class, symbolized as:
\begin{equation}
\exists y_{\text{common}}:|\{i :y^*_{\mathcal{SP},i} = y_{\text{common}}\}| = 8.
\end{equation}

\subsubsection{Dichotomy in Prediction for Trigger-Embedded Point Clouds.} Conversely, if the original point cloud \( P = \mathcal{B}(P';V) \) containing a trigger \( V \), where \( P' \) follows the distribution \( H_s \), the prediction outcomes for the sub-point clouds are expected to bifurcate into two categories within each group. Specifically, one category would encompass seven sub-point clouds converging to target class (designated by the attacker through the trigger mechanism), while the remaining sub-point cloud aligns with a different class, typically representing the original, correct classification, also known as the source class. This dichotomy in the predictions highlights the distinct behavior induced by the presence of the backdoor trigger within the point cloud data. Formally, this can be expressed as:
\begin{align}
    \exists y_{\text{target}}, y_{\text{source}} : 
   &|\{i : y^*_{\mathcal{SP},i} = y_{\text{target}}\}| = 7 \text{ and }\\ &|\{i : y^*_{\mathcal{SP},i} = y_{\text{source}}\}| = 1.\nonumber
\end{align}

These principles underpin \( \mathcal{DF} \)'s ability to discern the true classification of \( P \) by analyzing the pattern of predictions across the sub-point clouds. This approach effectively leverages the ensemble's collective intelligence to counteract the potential influence of a backdoor trigger, thereby enhancing the robustness and reliability of the classification process in the presence of adversarial manipulations.

But the aforementioned principles represent idealized scenarios in the CloudFort framework's functioning. In practical experimental settings, the application of these principles often encounters challenges stemming from the initial spatial partitioning step (Step 1). These challenges can impact the consistency and dichotomy of predictions, as outlined in the principles. Let's delve into how these challenges affect the process:

\subsubsection{Impact of Challenge 1 (Potential Removal of Critical Features) on Principles.} The first challenge in spatial partitioning is the potential removal of critical features from the point cloud, which could lead to a reduction in prediction consistency for non-triggered point clouds and affect the clear dichotomy in predictions for triggered point clouds. If essential features are lost in one or more sub-point clouds, the uniformity expected in the predictions for a non-triggered point cloud may not be as pronounced, leading to:

   \begin{equation}
   \exists y_{\text{common}} : |\{i : y^*_{\mathcal{SP},i} = y_{\text{common}}\}| < 8.
   \end{equation}

Similarly, for a triggered point cloud, the expected 7:1 ratio in the dichotomy of predictions might not manifest as distinctly if critical features impacting the trigger's detection are omitted.

\subsubsection{Impact of Challenge 2 (Introduction of Inherent Backdoors) on Principles.} The second challenge is the inadvertent introduction of configurations that might act as inherent backdoors during the partitioning process. This challenge can skew the expected outcomes, particularly for the triggered point clouds. Instead of observing a clear division in the predictions, the introduced backdoors might cause an unpredictable spread in the prediction outcomes:

\begin{align}
    \exists y_{\text{target}}, y_{\text{source}} : 
   &|\{i : y^*_{\mathcal{SP},i} = y_{\text{target}}\}| < 7 \text{ and }\\ &|\{i : y^*_{\mathcal{SP},i} = y_{\text{source}}\}| \neq 1.\nonumber
\end{align}

Similarly, for a non-triggered point cloud, the scenario is inherently nuanced due to the potential influence of inherent backdoors that might arise during the spatial partitioning process.

While the principles provide a theoretical framework for understanding CloudForts' operation, the real-world application is nuanced and influenced by the inherent challenges in spatial partitioning. These challenges necessitate a robust analysis techniques to ensure the effectiveness of CloudFort in discerning true classifications amidst potential adversarial manipulations.

Building upon the two aforementioned principles and the associated challenges, the process of determining the final \(y_{true}\) can be approached in a two-step manner. The first step involves analyzing the patterns exhibited by \(\mathcal{F}_\mathcal{V}(\Sigma_{CloudFort}(P))\) to ascertain whether the original input point cloud \(P\) carries a backdoor trigger. The second step leverages the findings from the first step to determine the ultimate value of \(y_{true}\).

\subsubsection{Trigger Presence Determination.}

To determine the presence of a trigger in the original point cloud \(P\), we introduce a function \(\mathcal{T}\) that evaluates the prediction patterns across the sub-point clouds generated by \(\Sigma_{CloudFort}\). Formally, let \(\mathcal{L} = \{y^*_{\mathcal{SP},i} : \mathcal{SP} \in \Sigma_{CloudFort}, i \in \{1, 2, \ldots, 8\}\}\) be the set of all predicted labels from \(\mathcal{F}_\mathcal{V}(\Sigma_{CloudFort}(P))\):

\begin{equation}
\mathcal{T}(\mathcal{L}) \rightarrow \{0, 1\},
\end{equation}

where an output of 0 signifies the absence of a trigger, and 1 indicates its presence. The function \(\mathcal{T}\) operates by examining the consistency and dichotomy of the labels within each partitioning strategy group \(\mathcal{SP}\).

Let \(\mathcal{C}_\mathcal{SP}\) be the set of unique labels predicted within a partitioning strategy group \(\mathcal{SP}\), and \(|\mathcal{C}_\mathcal{SP}|\) denote its cardinality. The number of partitioning strategies with \(|\mathcal{C}_\mathcal{SP}| \geq 2\), denoted as \(\gamma = \sum_{\mathcal{SP} \in \Sigma_{CloudFort}} \mathbf{1}_{|\mathcal{C}_\mathcal{SP}| \geq 2}\), where \(\mathbf{1}_{|\mathcal{C}_\mathcal{SP}| \geq 2}\) is an indicator function that equals 1 if the condition \(|\mathcal{C}_\mathcal{SP}| \geq 2\) is satisfied, and 0 otherwise. The total number of unique predicted labels across all sub-point clouds, denoted as \(\delta = |\bigcup_{\mathcal{SP} \in \Sigma_{CloudFort}} \mathcal{C}_\mathcal{SP}|\). Considering the challenges of potential removal of critical features and introduction of inherent backdoors, \(\mathcal{T}\) employs a relaxed criterion for trigger presence determination. Instead of strictly adhering to the idealized principles, it allows for some deviations. The trigger presence is determined as follows:

\begin{equation}
\mathcal{T}(\mathcal{L}) = 
\begin{cases}
    0, &\gamma \leq 2 \\
    0, &\gamma = 3 \text{ and } \delta \geq 4 \\
    0, &\gamma = 4 \text{ and } \delta \geq 5 \\
    1, &\gamma = 3 \text{ and } \delta < 4 \\
    1, &\gamma = 4 \text{ and } \delta < 5
\end{cases}
\end{equation}

\(\gamma\) reflects the inconsistency of predicted labels across different partitioning strategies, and \(\delta\) reflects the diversity of the prediction results. By combining different values of \(\gamma\) and \(\delta\), and considering the challenges posed by spatial partitioning, the \(\mathcal{T}\) function adopts a relaxed criterion for determining the presence of a trigger. The conditions for determining the presence of a trigger are carefully designed to align with the principles and challenges discussed in the paper. When \(\gamma \leq 2\), it indicates that the predicted labels are consistent across most partitioning strategies, which aligns with the expectation of a trigger-free scenario, thus concluding the absence of a trigger. When \(\gamma = 3\) and \(\delta \geq 4\), or \(\gamma = 4\) and \(\delta \geq 5\), despite the presence of some inconsistency, the high value of \(\delta\) suggests sufficient diversity in the prediction results. This could be attributed to the challenges brought about by spatial partitioning, such as the removal of critical features or the introduction of inherent backdoors, rather than the influence of a trigger, thus concluding the absence of a trigger. Conversely, when \(\gamma = 3\) and \(\delta < 4\), or \(\gamma = 4\) and \(\delta < 5\), there exists a high level of inconsistency and insufficient diversity in the prediction results, which could be indicative of the presence of a trigger, thus concluding the presence of a trigger.

\subsubsection{True Label Determination.}

Define a function \(\mathcal{CNT}: \mathcal{L} \rightarrow \mathbb{N}\) that maps each label \(y \in \mathcal{L}\) to its frequency of occurrence within \(\mathcal{L}\):

\begin{equation}
\mathcal{CNT}(y) = |\{(\mathcal{SP}, i) : y^*_{\mathcal{SP},i} = y\}|, \mathcal{SP}\in\Sigma_{CloudFort}.
\end{equation}
where \(|\cdot|\) denotes the cardinality of a set.

With the trigger presence information obtained from Step 1, the true label \(y_{true}\) can be determined using the decision function \(\mathcal{DF}\):

\begin{equation}
\mathcal{DF}(\mathcal{F}_\mathcal{V}(\Sigma_{CloudFort}(P))) \rightarrow \mathcal{DF}(\mathcal{L}, \mathcal{T}(\mathcal{L})) \rightarrow y_{true}.
\end{equation}

The function \(\mathcal{DF}\) takes into account both the prediction outcomes and the trigger presence information to make the final classification decision:

\begin{equation}
\mathcal{DF}(\mathcal{L}, \mathcal{T}(\mathcal{L}))=
\begin{cases}
\mathcal{F}_\mathcal{V}(P), &\mathcal{T}(\mathcal{L}) = 0 \\
\argmin_{y \in \mathcal{L}}|\mathcal{CNT}(y)-4|, & \mathcal{T}(\mathcal{L}) = 1.
\end{cases}
\end{equation}

When \(\mathcal{T}(\mathcal{L}) = 0\), indicating the absence of a trigger in the original point cloud \(P\), \(\mathcal{DF}\) directly utilizes the prediction result of the compromised classifier \(\mathcal{F}_\mathcal{V}\) on the original point cloud \(P\). This aligns with the principle of consistency in predictions for non-triggered point clouds, as the absence of a trigger suggests that the compromised classifier's prediction on the original point cloud is likely to be correct. On the other hand, when \(\mathcal{T}(\mathcal{L}) = 1\), signifying the presence of a trigger, \(\mathcal{DF}\) employs a more nuanced approach to determine the true label. In this case, \(\mathcal{DF}\) selects the label \(y \in \mathcal{L}\) that minimizes the absolute difference between its frequency of occurrence \(\mathcal{CNT}(y)\) and the value 4. This approach is grounded in the principle of dichotomy in predictions for trigger-embedded point clouds, which suggests that the true label is likely to be the one that appears less frequently among the sub-point clouds. The rationale behind this approach is that, in the presence of a trigger, the compromised classifier is expected to predict the target label for the majority of the sub-point clouds (i.e., 7 out of 8), while the true label is likely to be predicted for only a small subset of the sub-point clouds (i.e., 1 out of 8). By selecting the label that appears closest to 4 times, \(\mathcal{DF}\) aims to identify the true label amidst the triggered predictions.

Together, the spatial partitioning and ensemble prediction modules constitute the core defense pipeline of CloudFort, strategically constructed to offer protection against backdoor attacks on 3D point cloud classification systems.

\section{Experimental Evaluation}
\label{sec:experiment}
\subsection{Experimental Setup}

\subsubsection{Victim classification models.} To validate the effectiveness of the proposed CloudFort defense strategy, we conduct extensive experiments on three state-of-the-art point cloud classification models as victim classification models: PointNet~\cite{qi2017pointnet}, PointNet++~\cite{qi2017pointnet++}, and DGCNN~\cite{wang2019dynamic}. These models are selected due to their widespread adoption and strong performance in various point cloud classification tasks.

\subsubsection{Dataset.} The experiments are conducted on the ModelNet40 dataset, a widely used benchmark for point cloud classification\cite{wu20153d}. ModelNet40 contains 12,311 CAD models from 40 object categories, with 9,843 models used for training and 2,468 models used for testing. Each model is sampled to 1,024 points, normalized to a unit sphere, and augmented by random rotation and jittering, following the same data preprocessing strategy as PCBA~\cite{xiang2021backdoor}.

\subsubsection{Attack settings.} To evaluate the effectiveness of CloudFort against PCBA, we follow the attack methodology described in \cite{xiang2021backdoor} to generate backdoored versions of victim classification models. The attacker's goal is to misclassify a specific source class into a target class by inserting a backdoor trigger, which is a small cluster of points, into the point cloud. The backdoor triggers are designed to be stealthy and can effectively manipulate the predictions of the compromised models when present in the input point clouds. The backdoor trigger is designed as a cluster of 32 points, and its spatial location is optimized using the approach proposed in the attack paper. In this experiment, we adopt the RS (Random Sphere) local geometry from PCBA, as the spatial location of the trigger is the most critical factor for the success of the attack, while the local geometry is mainly used to counter anomaly detection. The poisoned training set and poisoned testing set are created following the methodology specified in the PCBA paper\cite{xiang2021backdoor}. Poisoned training set is used to train the victim classification models, aiming to inject the backdoor functionality into the models. Poisoned testing set is used to evaluate the attack success rate and the model's vulnerability to the backdoor trigger when presented with samples from the source class. By simulating the PCBA under these attack settings, we aim to evaluate the effectiveness of CloudFort in defending against backdoor attacks in point cloud classification models. 

We randomly select 8 pairs of source and target classes from ModelNet40 to generate different attack scenarios, which are: (laptop, chair), (cone, lamp), (chair, toilet), (keyboard, stair), (bed, glassbox), (bottle, car), (car, plant), (person, car).  To simplify the representation of the source and target class pairs, we denote them as P1, P2, ..., P8, respectively. The defense results will demonstrate the robustness of CloudFort in mitigating the impact of the PCBA and maintaining the integrity of the classification results.

\subsubsection{Performance evaluation metrics.} 

\paragraph{ACC (Clean test accuracy).} The classification accuracy on the clean test set without backdoor triggers.

\paragraph{ASR (Attack success rate).} The percentage of triggered test set that are classified into the target class. 

\paragraph{SIA (Source inference accuracy).} The percentage of triggered test set that are correctly classified back to their original source class. 

A successful backdoor attack is characterized by a high attack success rate (ASR), indicating that the backdoored model consistently misclassifies triggered test samples into the target class. Simultaneously, the attack should maintain a high clean test accuracy (ACC), ensuring that the model's performance on benign samples is not significantly degraded. Conversely, the source inference accuracy (SIA), which measures the model's ability to correctly classify triggered samples back to their original source class, should be low, reflecting the effectiveness of the backdoor in overriding the model's original decision-making process.

On the other hand, an effective mitigation approach aims to neutralize the backdoor's influence and restore the model's original behavior. A successful defense should significantly reduce the ASR, demonstrating that the backdoor's ability to manipulate the model's predictions has been suppressed. Meanwhile, the ACC should remain high, indicating that the defense does not compromise the model's performance on clean samples. Lastly, a high SIA is desired, as it signifies that the mitigation approach has successfully recovered the model's capability to correctly classify triggered samples back to their true source class, effectively eliminating the backdoor's malicious functionality.

All experiments are conducted on a GPU server with an NVIDIA Tesla V100S-PCIE-32GB GPU and implemented using the PyTorch deep learning framework\cite{paszke2019pytorch}.

\begin{table}[!ht]
\centering
\caption{Experimental results of CloudFort against PCBA on different models (in percentage \%).}
\label{tab:results}
\begin{tabular}{lccccccccc}
\hline
Model & Metric & P1 & P2 & P3 & P4 & P5 & P6 & P7 & P8 \\
\hline
\multirow{3}{*}{PointNet} & ASR & 0.0 & 5.0 & 14.0 & 0.0 & 9.0 & 0.0 & 94.0 & 10.0 \\
& ACC & 81.3 & 82.3 & 81.0 & 80.1 & 79.4 & 81.8 & 80.8 & 82.3 \\
& SIA & 100.0 & 80.0 & 79.0 & 95.0 & 82.0 & 95.0 & 2.0 & 75.0 \\
\hline
\multirow{3}{*}{PointNet++} & ASR & 0.0 & 5.0 & 1.0 & 0.0 & 4.0 & 0.0 & 76.0 & 5.0 \\
& ACC & 84.7 & 84& 84.6 & 84.4 & 83.9 & 84.8 & 84.2 & 84.1 \\
& SIA & 100.0 & 85.0 & 93.0 & 100.0 & 93.0 & 97.0 & 24.0 & 85.0 \\
\hline
\multirow{3}{*}{DGCNN} & ASR & 5.0 & 1.0 & 5.0 & 5.0 & 9.0 & 3.0 & 63.0 & 0.0 \\
& ACC & 78.9 & 79.9 & 80.4 & 79.6 & 80.4 & 81.2 & 79.9 & 79.0\\
& SIA & 95.0 & 80.0 & 75.0 & 95.0 & 84.0 & 95.0 & 12.0 & 80.0 \\
\hline
\end{tabular}
\end{table}

\begin{figure}[!h]
\centering
\begin{subfigure}[b]{0.32\textwidth}
\includegraphics[width=\textwidth]{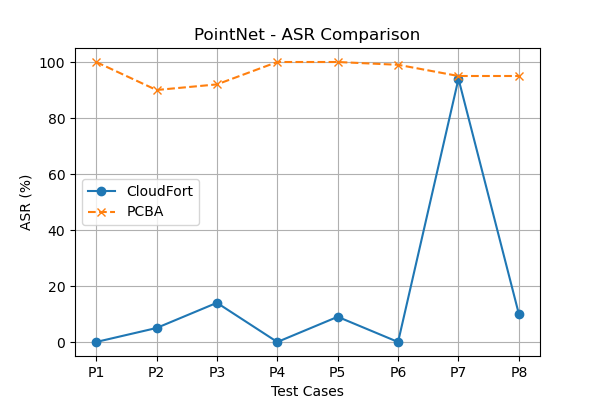}
\caption{Attack Success Rate (ASR) Comparison on PointNet}
\label{fig:asr of cloudfort and pcba on pointnet}
\end{subfigure}
\hfill
\begin{subfigure}[b]{0.32\textwidth}
\includegraphics[width=\textwidth]{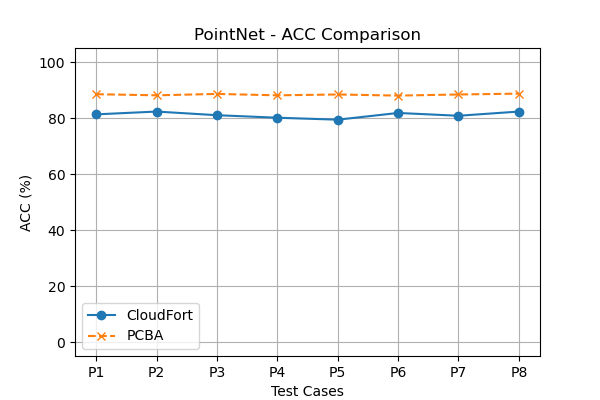}
\caption{Clean Test Accuracy (ACC) Comparison on PointNet}
\label{fig:acc of cloudfort and pcba on pointnet}
\end{subfigure}
\hfill
\begin{subfigure}[b]{0.32\textwidth}
\includegraphics[width=\textwidth]{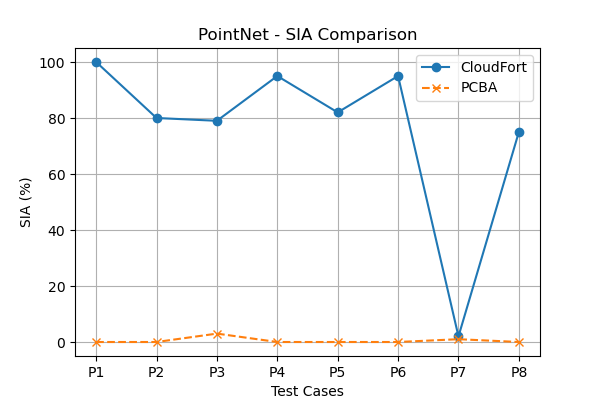}
\caption{Source Inference Accuracy (SIA) Comparison on PointNet}
\label{fig:sia of cloudfort and pcba on pointnet}
\end{subfigure}
\caption{Comparative Analysis of CloudFort and PCBA on PointNet}
\label{fig:comparison of cloudfort and pcba on pointnet}
\end{figure}

\begin{figure}[!h]
\centering
\begin{subfigure}[b]{0.32\textwidth}
\includegraphics[width=\textwidth]{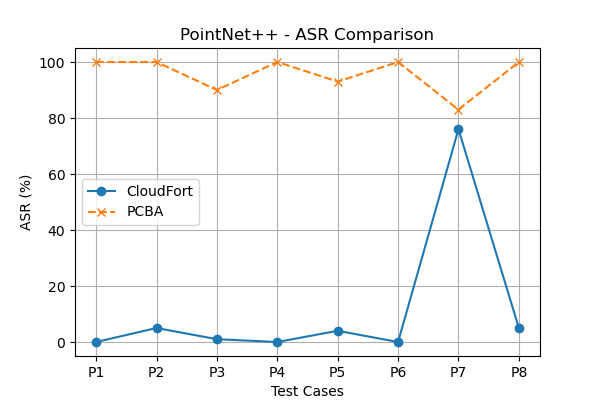}
\caption{Attack Success Rate (ASR) Comparison on PointNet++}
\label{fig:asr of cloudfort and pcba on pointnet++}
\end{subfigure}
\hfill
\begin{subfigure}[b]{0.32\textwidth}
\includegraphics[width=\textwidth]{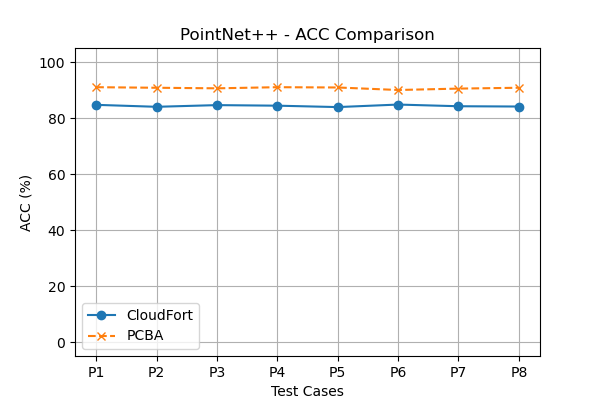}
\caption{Clean Test Accuracy (ACC) Comparison on PointNet++}
\label{fig:acc of cloudfort and pcba on pointnet++}
\end{subfigure}
\hfill
\begin{subfigure}[b]{0.32\textwidth}
\includegraphics[width=\textwidth]{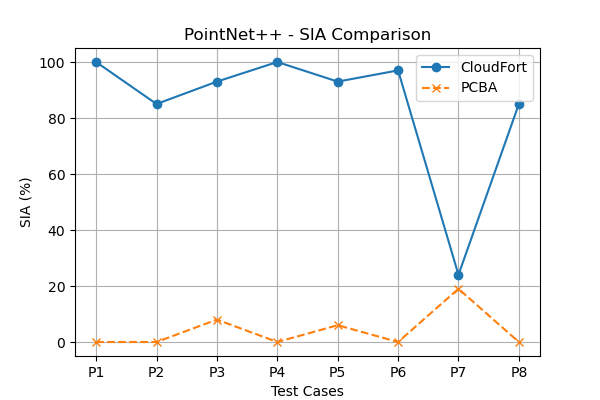}
\caption{Source Inference Accuracy (SIA) Comparison on PointNet++}
\label{fig:sia of cloudfort and pcba on pointnet++}
\end{subfigure}
\caption{Comparative Analysis of CloudFort and PCBA on PointNet++}
\label{fig:comparison of cloudfort and pcba on pointnet++}
\end{figure}

\begin{figure}[!h]
\centering
\begin{subfigure}[b]{0.32\textwidth}
\includegraphics[width=\textwidth]{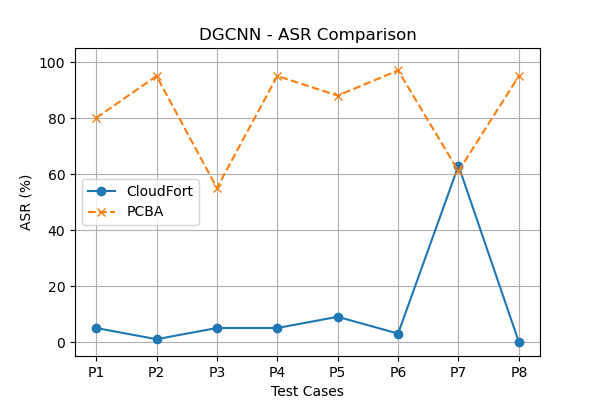}
\caption{Attack Success Rate (ASR) Comparison on DGCNN}
\label{fig:asr of cloudfort and pcba on dgcnn}
\end{subfigure}
\hfill
\begin{subfigure}[b]{0.32\textwidth}
\includegraphics[width=\textwidth]{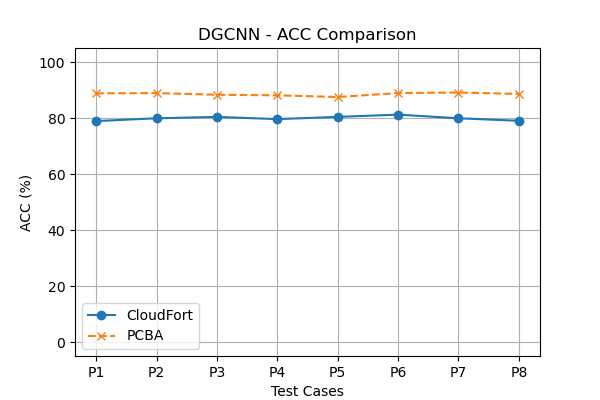}
\caption{Clean Test Accuracy (ACC) Comparison on DGCNN}
\label{fig:acc of cloudfort and pcba on dgcnn}
\end{subfigure}
\hfill
\begin{subfigure}[b]{0.32\textwidth}
\includegraphics[width=\textwidth]{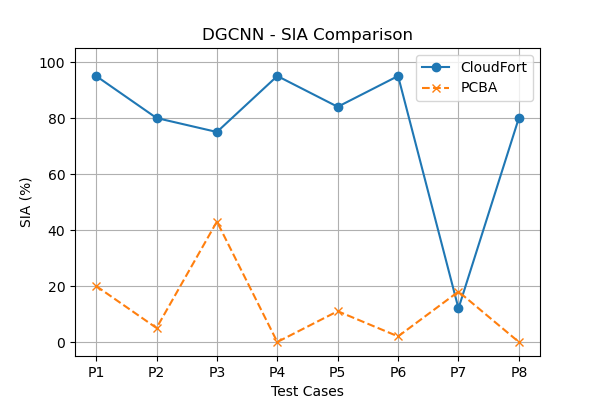}
\caption{Source Inference Accuracy (SIA) Comparison on DGCNN}
\label{fig:sia of cloudfort and pcba on dgcnn}
\end{subfigure}
\caption{Comparative Analysis of CloudFort and PCBA on DGCNN}
\label{fig:comparison of cloudfort and pcba on dgcnn}
\end{figure}

\subsection{Results and Analysis}

The experimental results presented in Table \ref{tab:results} and the three groups of comparison plots (Figure  \ref{fig:comparison of cloudfort and pcba on pointnet}, Figure \ref{fig:comparison of cloudfort and pcba on pointnet++} and Figure \ref{fig:comparison of cloudfort and pcba on dgcnn}) collectively demonstrate the effectiveness of CloudFort in mitigating the impact of backdoor attacks across different point cloud classification models.  Overall, CloudFort exhibits a strong defensive capability across various attack scenarios, successfully reducing the attack success rate (ASR) while maintaining high clean test accuracy (ACC) and restoring the model's ability to correctly classify triggered samples back to their original source class, as evidenced by the improved source inference accuracy (SIA).

Figure \ref{fig:asr of cloudfort and pcba on pointnet}, Figure \ref{fig:asr of cloudfort and pcba on pointnet++} and Figure \ref{fig:asr of cloudfort and pcba on dgcnn} illustrate the significant reduction in ASR achieved by CloudFort compared to PCBA across most test cases. This observation aligns with the results in Table \ref{tab:results}. In the majority of the evaluated scenarios (P1, P2, P3, P4, P5, P6, and P8), CloudFort effectively neutralizes the backdoor's influence, achieving ASRs below 15\% for all three models. Notably, in several cases, such as P1 and P6 for PointNet and PointNet++, and P8 for DGCNN, CloudFort completely eliminates the backdoor's functionality, reducing the ASR to 0\%. These results highlight the robustness of the proposed defense mechanism in successfully mitigating the impact of PCBA across different model architectures and attack configurations.

Moreover, Figure \ref{fig:acc of cloudfort and pcba on pointnet}, Figure \ref{fig:acc of cloudfort and pcba on pointnet++} and Figure \ref{fig:acc of cloudfort and pcba on dgcnn} show that CloudFort maintains relatively high ACC values, ranging from 78.9\% to 84.8\% across all models and attack scenarios, as also indicated in Table \ref{tab:results}. This indicates that CloudFort's defensive operations do not significantly compromise the model's performance on clean, benign samples. The preservation of high ACC values underscores the practicality of CloudFort, as it maintains the model's utility for its intended classification task while providing effective protection against backdoor attacks.

Furthermore, Figure \ref{fig:sia of cloudfort and pcba on pointnet}, Figure \ref{fig:sia of cloudfort and pcba on pointnet++} and Figure \ref{fig:sia of cloudfort and pcba on dgcnn} highlight CloudFort's superiority in terms of SIA. The SIA values demonstrate CloudFort's ability to recover the model's original decision-making process for triggered samples. In most cases, the SIA exceeds 75\%, reaching 100\% in some instances (e.g., P1 for PointNet and PointNet++, P4 for PointNet++), as supported by the corresponding entries in Table \ref{tab:results}. These results suggest that CloudFort successfully removes the backdoor's malicious functionality, enabling the model to correctly classify triggered samples based on their true source class rather than the attacker's target class.

However, it is important to acknowledge that CloudFort's defensive capabilities are not infallible, as evidenced by the failure case in P7. For this specific attack scenario, CloudFort struggles to effectively mitigate the backdoor's impact, resulting in high ASRs of 94\%, 76\%, and 63\% for PointNet, PointNet++, and DGCNN, respectively. Correspondingly, the SIA values for P7 are significantly lower compared to other scenarios, ranging from 2\% to 24\%. This indicates that the backdoor's influence persists, and the model continues to misclassify triggered samples into the attacker's target class despite CloudFort's defensive measures.

In section \ref{Analysis of CloudFort's Limitation}, we will delve into a comprehensive analysis of CloudFort's limitations, focusing specifically on the P7 scenario. By investigating the factors contributing to the defense failure, we aim to provide a deeper understanding of the challenges faced by CloudFort.

\begin{table}[!ht]
\centering
\caption{Experimental results of simplified CloudFort against PCBA on different models (in percentage \%).}
\label{tab:updated_results}
\begin{tabular}{lcccccccc}
\hline
Model & Metric & P1 & P2 & P3 & P4 & P5 & P6 & P8 \\
\hline
\multirow{3}{*}{PointNet} & ASR & 0.0 & 5.0 & 8.0 & 0.0 & 1.0 & 0.0 & 5.0 \\
& ACC & 78.9 & 79.9 & 76.9 & 72.1 & 76.5 & 78.7 & 78.8 \\
& SIA & 100.0 & 80.0 & 83.0 & 100.0 & 93.0 & 93.0 & 80.0 \\
\hline
\multirow{3}{*}{PointNet++} & ASR & 0.0 & 0.0 & 4.0 & 0.0 & 6.0 & 0.0 & 5.0 \\
& ACC & 81.3 & 82.1 & 82.4 & 81.8 & 82.1 & 81.5 & 81.8 \\
& SIA & 95.0 & 85.0 & 91.0 & 100.0 & 88.0 & 97.0 & 85.0 \\
\hline
\multirow{3}{*}{DGCNN} & ASR & 10.0 & 15.0 & 14.0 & 5.0 & 9.0 & 1.0 & 0.0 \\
& ACC & 75.2 & 76.1 & 78.4 & 77.3 & 77.1 & 77.0 & 75.8 \\
& SIA & 90.0 & 80.0 & 77.0 & 95.0 & 86.0 & 96.0 & 85.0 \\
\hline
\end{tabular}
\end{table}

\begin{figure}[!ht]
\centering
\begin{subfigure}[b]{0.32\textwidth}
\includegraphics[width=\textwidth]{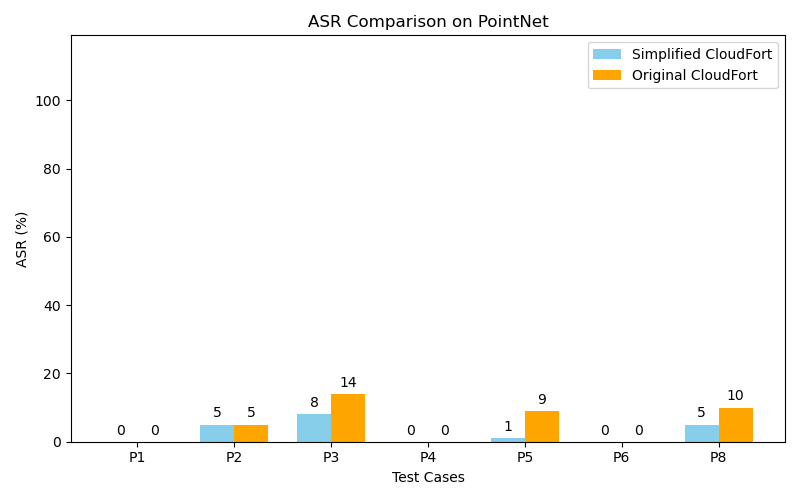}
\caption{ASR Comparison on PointNet}
\label{fig:asr of Simplified and Original CloudFort on pointnet}
\end{subfigure}
\hfill
\begin{subfigure}[b]{0.32\textwidth}
\includegraphics[width=\textwidth]{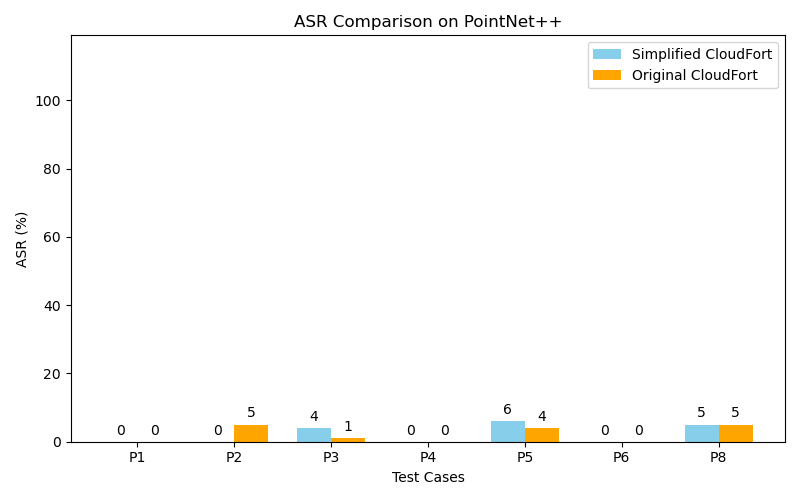}
\caption{ASR Comparison on PointNet++}
\label{fig:asr of cloudfort and simplified cloudfort on pointnet++}
\end{subfigure}
\hfill
\begin{subfigure}[b]{0.32\textwidth}
\includegraphics[width=\textwidth]{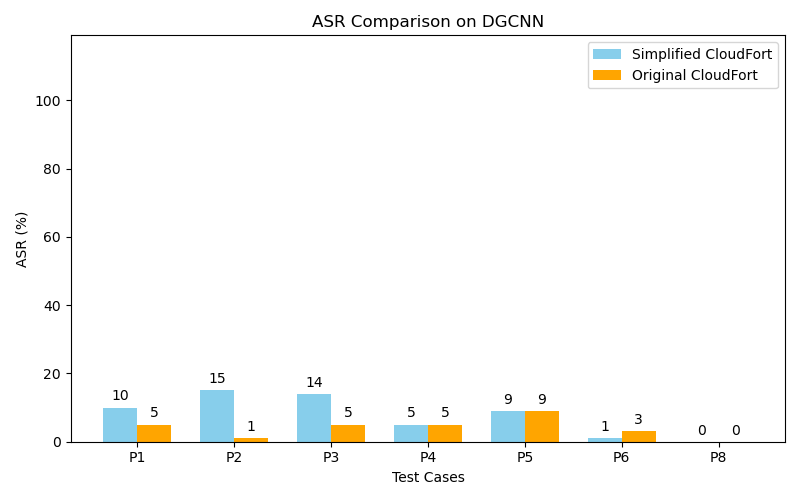}
\caption{ASR Comparison on DGCNN}
\label{fig:asr of cloudfort and simplified cloudfort on dgcnn}
\end{subfigure}
\caption{ASR Comparison of Simplified and Original CloudFort}
\label{fig:asr comparison of Simplified and Original CloudFort}
\end{figure}

\begin{figure}[!ht]
\centering
\begin{subfigure}[b]{0.32\textwidth}
\includegraphics[width=\textwidth]{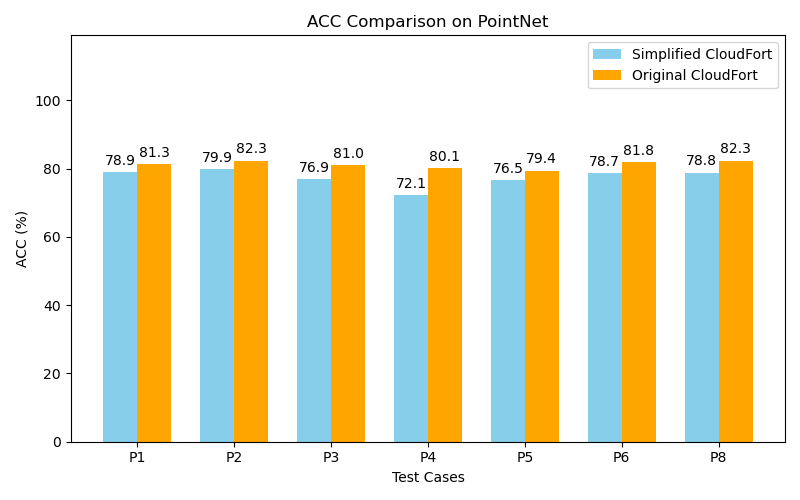}
\caption{ACC Comparison on PointNet}
\label{fig:acc of cloudfort and simplified cloudfort on pointnet}
\end{subfigure}
\hfill
\begin{subfigure}[b]{0.32\textwidth}
\includegraphics[width=\textwidth]{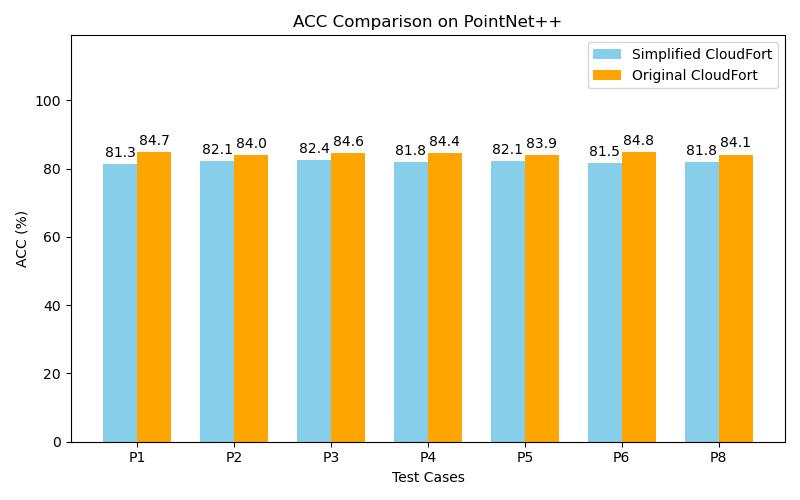}
\caption{ACC Comparison on PointNet++}
\label{fig:acc of cloudfort and simplified cloudfort on pointnet++}
\end{subfigure}
\hfill
\begin{subfigure}[b]{0.32\textwidth}
\includegraphics[width=\textwidth]{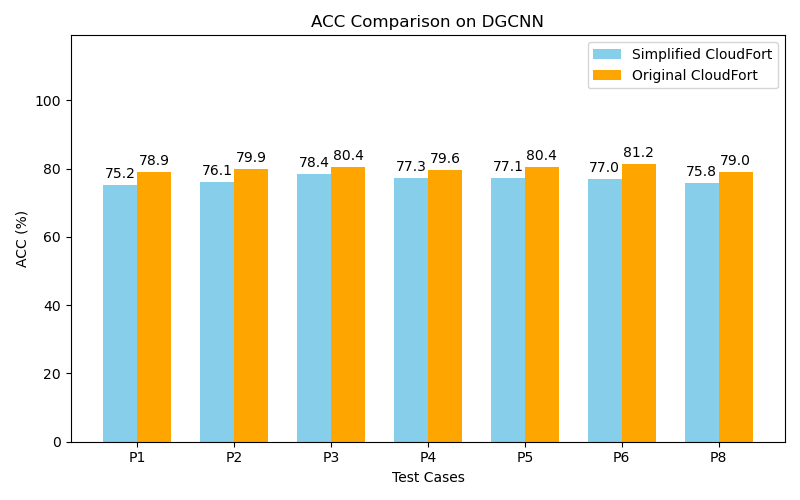}
\caption{ACC Comparison on DGCNN}
\label{fig:acc of cloudfort and simplified cloudfort on dgcnn}
\end{subfigure}
\caption{ACC Comparison of Simplified and Original CloudFort}
\label{fig:acc comparison of Simplified and Original CloudFort}
\end{figure}

\begin{figure}[!ht]
\centering
\begin{subfigure}[b]{0.32\textwidth}
\includegraphics[width=\textwidth]{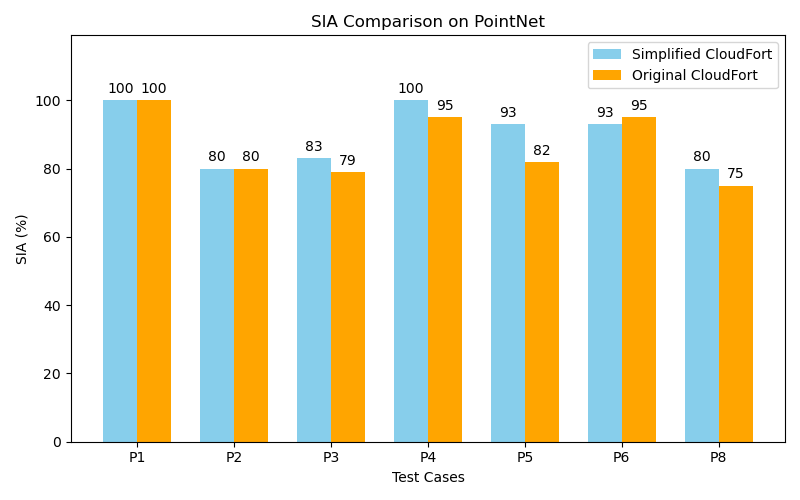}
\caption{SIA Comparison on PointNet}
\label{fig:sia of cloudfort and simplified cloudfort on pointnet}
\end{subfigure}
\hfill
\begin{subfigure}[b]{0.32\textwidth}
\includegraphics[width=\textwidth]{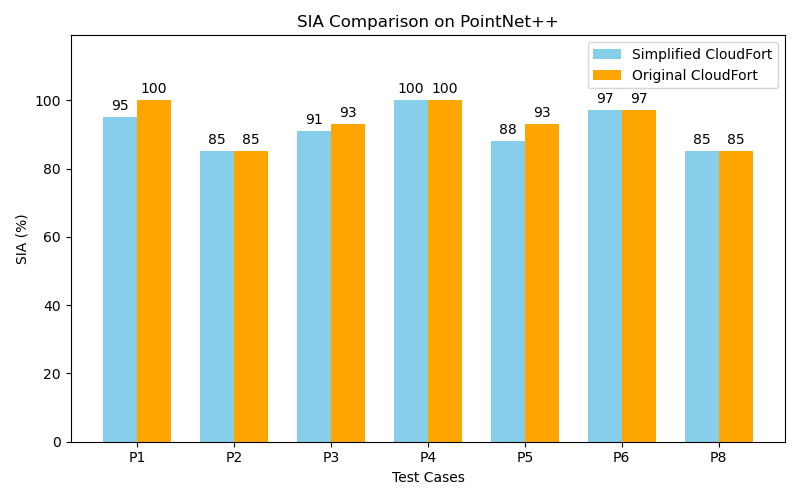}
\caption{SIA Comparison on PointNet++}
\label{fig:sia of cloudfort and simplified cloudfort on pointnet++}
\end{subfigure}
\hfill
\begin{subfigure}[b]{0.32\textwidth}
\includegraphics[width=\textwidth]{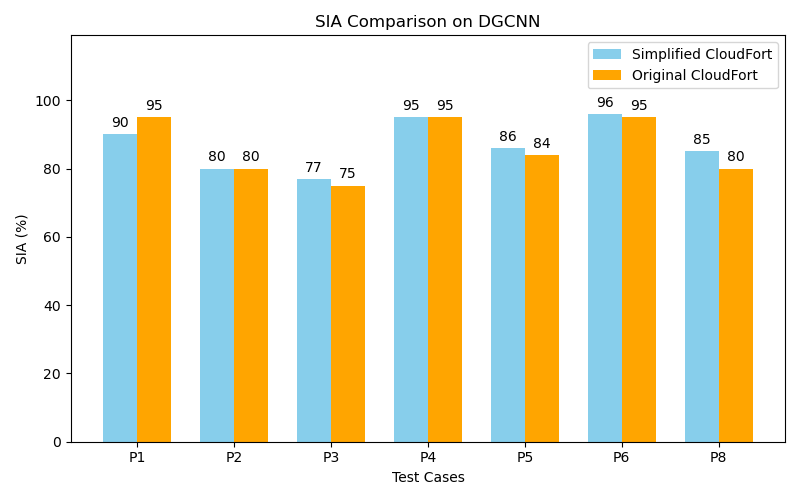}
\caption{SIA Comparison on DGCNN}
\label{fig:sia of cloudfort and simplified cloudfort on dgcnn}
\end{subfigure}
\caption{SIA Comparison of Simplified and Original CloudFort}
\label{fig:sia comparison of Simplified and Original CloudFort}
\end{figure}

\subsection{Ablation Study}

To evaluate the contribution of different components in CloudFort's defense mechanism, we conducted an ablation study by simplifying the spatial partitioning and ensemble prediction steps. In the simplified version of CloudFort, we applied only one spatial partitioning strategy, \( \Sigma_{CloudFort} = \{\mathcal{SP}_1\} \), which uses the standard partitioning without rotation. This results in a single group of 8 sub-point clouds. For the ensemble prediction step, we used simplified rules based on two principles: (1) Consistency in Prediction for Non-Triggered Point Clouds, and (2) Dichotomy in Prediction for Trigger-Embedded Point Clouds, as described in the previous section. These principles were used to determine the presence of a trigger and infer the true label.

Table \ref{tab:updated_results} presents the performance metrics of the simplified CloudFort against PCBA on PointNet, PointNet++, and DGCNN models. To compare the defense performance of the simplified and original versions of CloudFort, we focus on the results for test cases P1, P2, P3, P4, P5, P6, and P8.

Figure \ref{fig:asr comparison of Simplified and Original CloudFort} illustrates the ASR comparison between the simplified and original CloudFort for the selected test cases across all three models. The simplified CloudFort achieves comparable ASR results to the original version in most cases, with ASR values ranging from 0\% to 15\%. This indicates that even with a single spatial partitioning strategy and simplified ensemble prediction rules, CloudFort can effectively mitigate the impact of backdoor attacks in the majority of scenarios.

Figure \ref{fig:acc comparison of Simplified and Original CloudFort} compares the ACC values of the simplified and original CloudFort. The simplified version maintains high ACC values, ranging from 72.1\% to 82.4\% across all models and test cases. However, there is a slight decrease in ACC compared to the original CloudFort, particularly for the PointNet model. This suggests that the diversity of spatial partitioning strategies in the original CloudFort contributes to its ability to preserve the model's performance on benign samples.

Figure \ref{fig:sia comparison of Simplified and Original CloudFort} presents the SIA comparison between the simplified and original CloudFort. The simplified version achieves high SIA values, ranging from 77\% to 100\% across all models and test cases. These results are comparable to the original CloudFort, indicating that the simplified ensemble prediction rules are still effective in restoring the model's ability to correctly classify triggered samples back to their original source class.

The ablation study demonstrates that the simplified version of CloudFort, with a single spatial partitioning strategy and simplified ensemble prediction rules, can still provide effective defense against backdoor attacks in most scenarios. However, the slightly lower ACC values suggest that the diversity of spatial partitioning strategies in the original CloudFort contributes to its robustness in preserving the model's performance on benign samples.

It is worth noting that the simplified CloudFort's performance may be more sensitive to the choice of the spatial partitioning strategy, as it relies on a single approach. In contrast, the original CloudFort's use of multiple partitioning strategies with rotations helps to mitigate the potential limitations of any single strategy, such as the removal of critical features or the introduction of inherent backdoors.

Based on these findings, we can conclude that while the simplified CloudFort provides a viable defense solution, the additional components in the original CloudFort, particularly the diverse spatial partitioning strategies, contribute to its overall robustness and effectiveness. The original CloudFort's ability to maintain high ACC values while effectively mitigating backdoor attacks highlights the importance of these components in its defense mechanism.

\subsection{Limitation of CloudFort} \label{Analysis of CloudFort's Limitation}

\begin{table}[!ht]
\centering
\caption{Experimental results of CloudFort against PCBA on PointNet (in percentage \%).}
\label{tab:limitation_results}
\begin{tabular}{lcccccc}
\hline
Model & Metric & P7 & P9 & P10 \\
\hline
\multirow{3}{*}{PointNet} & ASR & 94.0 & 6.0 & 18.0 \\
& ACC & 80.8 & 81.8 & 81.6 \\
& SIA & 2.0 & 92.0 & 74.0 \\
\hline
\end{tabular}
\end{table}

\begin{figure}[!ht]
\centering
\begin{subfigure}[b]{0.32\textwidth}
\includegraphics[width=\textwidth]{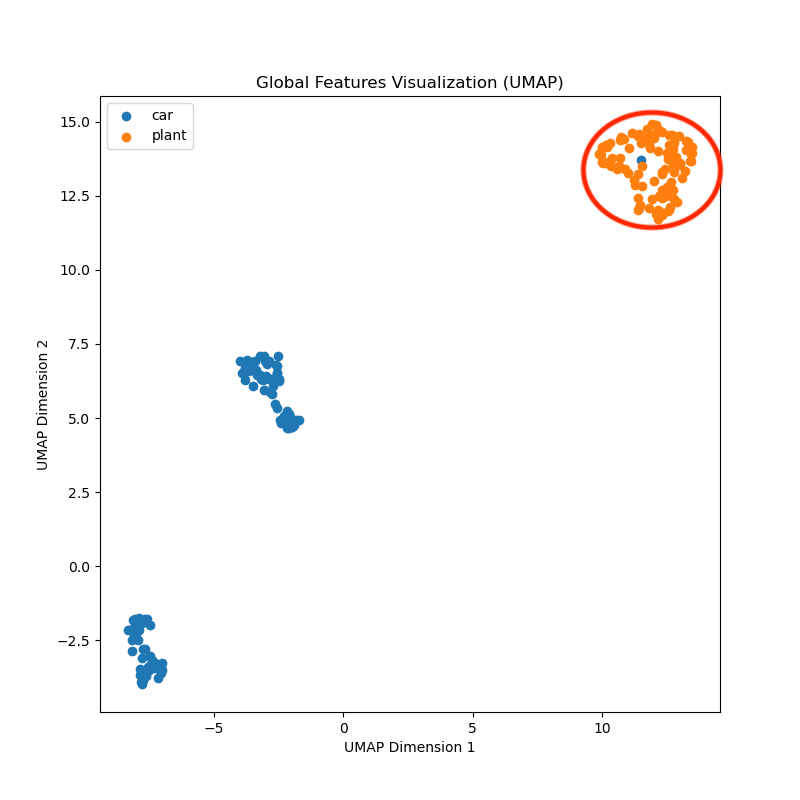}
\caption{P7 (car, plant)}
\label{fig:P7 visualization}
\end{subfigure}
\hfill
\begin{subfigure}[b]{0.32\textwidth}
\includegraphics[width=\textwidth]{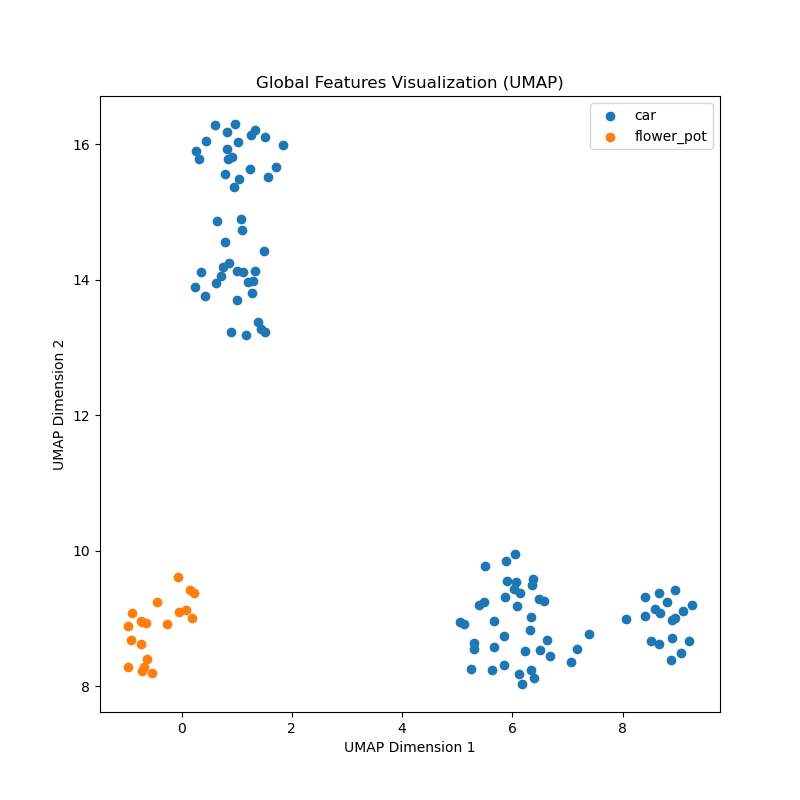}
\caption{P9 (car, flower\_pot)}
\label{fig:P9 visualization}
\end{subfigure} 
\hfill
\begin{subfigure}[b]{0.32\textwidth}
\includegraphics[width=\textwidth]{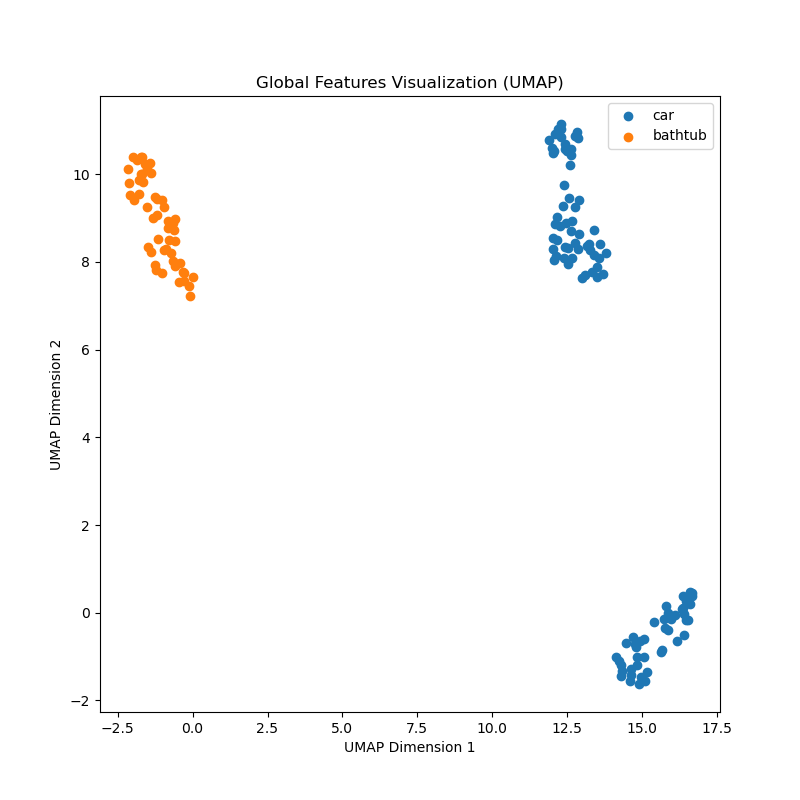}
\caption{P10 (car, bathtub)}
\label{fig:P10 visualization}
\end{subfigure}
\caption{Feature Representations Visualization}
\label{fig:visaulization of global features}
\end{figure}

The experimental results presented in Table \ref{tab:results} and the comparison plots highlight the effectiveness of CloudFort in mitigating the impact of PCBA across various attack scenarios. However, the P7 case, where the source class is "car" and the target class is "plant," stands out as a notable failure case for CloudFort's defense mechanism. In this scenario, CloudFort exhibits a significantly high attack success rate (ASR) of 94\% and a low source inference accuracy (SIA) of only 2\%, indicating its inability to effectively defend against the backdoor attack.

To investigate the potential reasons behind CloudFort's failure in the P7 scenario, we hypothesize that the inherent similarity between the source and target point clouds in the feature space may pose a challenge for the defense mechanism. If the "car" and "plant" classes share similar geometric characteristics, their feature representations learned by the compromised model might be relatively close to each other. As a result, even when the source point cloud is injected with a backdoor trigger, CloudFort's decision function \(\mathcal{DF}(\mathcal{L}, \mathcal{T}(\mathcal{L}))\) may struggle to distinguish between the feature changes caused by the trigger and the genuine similarities between the two classes.

To validate this hypothesis, we construct two additional attack scenarios: P9 (car, flower\_pot) and P10 (car, bathtub). These scenarios share the same source class as P7, which is "car," but have different target classes. By comparing the experimental results of CloudFort against PCBA on PointNet for P7, P9, and P10, we aim to gain insights into the impact of class similarity on CloudFort's defense performance.

Table \ref{tab:limitation_results} presents the experimental results for these three attack scenarios. In the P9 scenario, where the target class is "flower\_pot," CloudFort achieves a significantly lower ASR of 6\% and a higher SIA of 92\% compared to P7. Similarly, in the P10 scenario, where the target class is "bathtub," CloudFort exhibits an ASR of 18\% and an SIA of 74\%, which are still notably better than the results for P7.

To gain a deeper understanding of the similarity between the source and target classes in the P7, P9, and P10 scenarios, we extract samples from the ModelNet40 test set for the "car," "plant," "flower\_pot," and "bathtub" classes. We perform a visual analysis of their feature representations obtained from the compromised PointNet model. By employing the Uniform Manifold Approximation and Projection (UMAP) method ~\cite{mcinnes2018umap}, we map the high-dimensional features into a two-dimensional space, revealing the similarities and differences between the classes. Figure \ref{fig:visaulization of global features} visualizes their feature space representations .

Figure \ref{fig:P7 visualization} presents the feature visualization results for the "car" and "plant" classes in the P7 scenario. The plot clearly shows a significant overlap between the two classes in the learned UMAP space, with one data point from the "car" class falling in the midst of the "plant" class data points. This observation indicates a certain level of similarity between the "car" and "plant" classes in the feature space, aligning with the poor defense performance of CloudFort in the P7 scenario, as reported in Table \ref{tab:limitation_results}. When the source and target classes exhibit resemblance in the feature space, CloudFort's spatial partitioning and ensemble prediction steps may struggle to effectively distinguish between the feature changes caused by the backdoor trigger and the inherent similarities between the classes.

In contrast, Figure \ref{fig:P9 visualization} depicts the feature visualization results for the "car" and "flower\_pot" classes in the P9 scenario. Unlike P7, these two classes do not exhibit any overlap in the UMAP space, suggesting a lower degree of similarity. This observation is consistent with the improved defense performance of CloudFort in the P9 scenario, as evidenced by the lower ASR and higher SIA values in Table \ref{tab:limitation_results}.

Similarly, Figure \ref{fig:P10 visualization} illustrates the feature visualization results for the "car" and "bathtub" classes in the P10 scenario. Again, no overlap is observed between the two classes in the UMAP space, indicating a clear separation and dissimilarity in their feature representations. This finding aligns with the relatively better defense performance of CloudFort in the P10 scenario, as shown in Table \ref{tab:limitation_results}.

By comparing the feature visualization results of the source and target classes in the P7, P9, and P10 scenarios, we can draw the following observations:

1. A certain level of similarity between source and target classes in the feature space ,indeed impacts CloudFort's defense performance. When two classes exhibit a certain degree of similarity, as observed in the P7 scenario with the overlap between "car" and "plant" classes, CloudFort faces greater challenges in distinguishing the influence of the backdoor trigger from the inherent class similarities, leading to degraded defense performance.

2. The absence of overlap in the feature visualizations of P9 and P10 scenarios suggests that the source and target classes in these cases exhibit lower similarity compared to P7. This observation aligns with the improved defense performance of CloudFort in these scenarios, indicating that the reduced similarity enables CloudFort to more effectively isolate the impact of the backdoor trigger.

3. The clear separation and dissimilarity between the source and target classes in the feature space, as observed in the P10 scenario, provide more favorable conditions for CloudFort's defense mechanism. Although limitations may still exist, the distinct feature representations of the classes facilitate CloudFort's ability to differentiate between the feature changes induced by the backdoor trigger and the normal class features.

In conclusion, the feature visualization analysis of the source and target classes in the P7, P9, and P10 scenarios further validates the relationship between CloudFort's defense performance and class similarity. This analysis not only deepens our understanding of CloudFort's limitations but also provides valuable insights for improving the defense mechanism. Future work can explore methods to better incorporate class similarity considerations into the spatial partitioning and ensemble prediction processes, enhancing CloudFort's robustness in handling highly similar classes. Moreover, this analysis emphasizes the importance of considering the dynamics of class features in defending against backdoor attacks, opening new perspectives for ensuring the trustworthiness and security of point cloud classification models.

\section{Conclusions}
\label{sec:conclusion}

This study introduces CloudFort as a promising defense strategy against point cloud backdoor attacks(PCBA)\cite{xiang2021backdoor}, demonstrating its effectiveness through comprehensive experimental evaluations and comparative analyses. The insights gained from this research contribute to the broader understanding of the challenges and complexities involved in defending against backdoor attacks in point cloud classification tasks. By addressing the limitations and incorporating the lessons learned from this study, we can further enhance the robustness and generalizability of CloudFort and pave the way for more secure and trustworthy point cloud classification systems. As the field of adversarial machine learning continues to evolve, it is crucial to develop adaptive and resilient defense mechanisms that can keep pace with the ever-changing landscape of security threats in the realm of 3D point cloud processing.

\begin{credits}
\subsubsection{\ackname} This work is supported by the National Key R\&D Program of China (2022YFB4500402) and the Fundamental Research Funds for the Central Universities. The corresponding author of this paper is Haihua Shen.
\end{credits}

\bibliographystyle{splncs04}
\bibliography{cloudfort}

\begin{thebibliography}{10}
\providecommand{\url}[1]{\texttt{#1}}
\providecommand{\urlprefix}{URL }
\providecommand{\doi}[1]{https://doi.org/#1}

\bibitem{alexiou2020pointxr}
Alexiou, E., Yang, N., Ebrahimi, T.: Pointxr: A toolbox for visualization and
  subjective evaluation of point clouds in virtual reality. In: 2020 Twelfth
  International Conference on Quality of Multimedia Experience (QoMEX).
  pp.~1--6. IEEE (2020)

\bibitem{blanc2020genuage}
Blanc, T., El~Beheiry, M., Caporal, C., Masson, J.B., Hajj, B.: Genuage:
  visualize and analyze multidimensional single-molecule point cloud data in
  virtual reality. Nature Methods  \textbf{17}(11),  1100--1102 (2020)

\bibitem{borgnia2021strong}
Borgnia, E., Cherepanova, V., Fowl, L., Ghiasi, A., Geiping, J., Goldblum, M.,
  Goldstein, T., Gupta, A.: Strong data augmentation sanitizes poisoning and
  backdoor attacks without an accuracy tradeoff. In: ICASSP 2021-2021 IEEE
  International Conference on Acoustics, Speech and Signal Processing (ICASSP).
  pp. 3855--3859. IEEE (2021)

\bibitem{chen2022direct}
Chen, K., Lopez, B.T., Agha-mohammadi, A.a., Mehta, A.: Direct lidar odometry:
  Fast localization with dense point clouds. IEEE Robotics and Automation
  Letters  \textbf{7}(2),  2000--2007 (2022)

\bibitem{chen20203d}
Chen, S., Liu, B., Feng, C., Vallespi-Gonzalez, C., Wellington, C.: 3d point
  cloud processing and learning for autonomous driving: Impacting map creation,
  localization, and perception. IEEE Signal Processing Magazine
  \textbf{38}(1),  68--86 (2020)

\bibitem{cui2021deep}
Cui, Y., Chen, R., Chu, W., Chen, L., Tian, D., Li, Y., Cao, D.: Deep learning
  for image and point cloud fusion in autonomous driving: A review. IEEE
  Transactions on Intelligent Transportation Systems  \textbf{23}(2),  722--739
  (2021)

\bibitem{di2021mobile}
Di~Stefano, F., Chiappini, S., Gorreja, A., Balestra, M., Pierdicca, R.: Mobile
  3d scan lidar: A literature review. Geomatics, Natural Hazards and Risk
  \textbf{12}(1),  2387--2429 (2021)

\bibitem{doan2020februus}
Doan, B.G., Abbasnejad, E., Ranasinghe, D.C.: Februus: Input purification
  defense against trojan attacks on deep neural network systems. In:
  Proceedings of the 36th Annual Computer Security Applications Conference. pp.
  897--912 (2020)

\bibitem{dong2021black}
Dong, Y., Yang, X., Deng, Z., Pang, T., Xiao, Z., Su, H., Zhu, J.: Black-box
  detection of backdoor attacks with limited information and data. In:
  Proceedings of the IEEE/CVF International Conference on Computer Vision. pp.
  16482--16491 (2021)

\bibitem{duan2021robotics}
Duan, H., Wang, P., Huang, Y., Xu, G., Wei, W., Shen, X.: Robotics dexterous
  grasping: The methods based on point cloud and deep learning. Frontiers in
  Neurorobotics  \textbf{15},  658280 (2021)

\bibitem{gao2023imperceptible}
Gao, K., Bai, J., Wu, B., Ya, M., Xia, S.T.: Imperceptible and robust backdoor
  attack in 3d point cloud. IEEE Transactions on Information Forensics and
  Security  \textbf{19},  1267--1282 (2023)

\bibitem{gu2017badnets}
Gu, T., Dolan-Gavitt, B., Garg, S.: Badnets: Identifying vulnerabilities in the
  machine learning model supply chain. arXiv preprint arXiv:1708.06733  (2017)

\bibitem{hu2023pointcrt}
Hu, S., Liu, W., Li, M., Zhang, Y., Liu, X., Wang, X., Zhang, L.Y., Hou, J.:
  Pointcrt: Detecting backdoor in 3d point cloud via corruption robustness. In:
  Proceedings of the 31st ACM International Conference on Multimedia. pp.
  666--675 (2023)

\bibitem{lee2023robust}
Lee, M., Kim, D.: Robust evaluation of diffusion-based adversarial
  purification. In: Proceedings of the IEEE/CVF International Conference on
  Computer Vision. pp. 134--144 (2023)

\bibitem{li2021pointba}
Li, X., Chen, Z., Zhao, Y., Tong, Z., Zhao, Y., Lim, A., Zhou, J.T.: Pointba:
  Towards backdoor attacks in 3d point cloud. In: Proceedings of the IEEE/CVF
  international conference on computer vision. pp. 16492--16501 (2021)

\bibitem{li2024pointcvar}
Li, X., Lu, J., Ding, H., Sun, C., Zhou, J.T., Chee, Y.M.: Pointcvar:
  Risk-optimized outlier removal for robust 3d point cloud classification. In:
  Proceedings of the AAAI Conference on Artificial Intelligence. vol.~38, pp.
  21340--21348 (2024)

\bibitem{li2022backdoor}
Li, Y., Jiang, Y., Li, Z., Xia, S.T.: Backdoor learning: A survey. IEEE
  Transactions on Neural Networks and Learning Systems  (2022)

\bibitem{liu2021pointguard}
Liu, H., Jia, J., Gong, N.Z.: Pointguard: Provably robust 3d point cloud
  classification. In: Proceedings of the IEEE/CVF conference on computer vision
  and pattern recognition. pp. 6186--6195 (2021)

\bibitem{liu2020reflection}
Liu, Y., Ma, X., Bailey, J., Lu, F.: Reflection backdoor: A natural backdoor
  attack on deep neural networks. In: Computer Vision--ECCV 2020: 16th European
  Conference, Glasgow, UK, August 23--28, 2020, Proceedings, Part X 16. pp.
  182--199. Springer (2020)

\bibitem{maturana2018real}
Maturana, D., Chou, P.W., Uenoyama, M., Scherer, S.: Real-time semantic mapping
  for autonomous off-road navigation. In: Field and Service Robotics: Results
  of the 11th International Conference. pp. 335--350. Springer (2018)

\bibitem{mcinnes2018umap}
McInnes, L., Healy, J., Melville, J.: Umap: Uniform manifold approximation and
  projection for dimension reduction. arXiv preprint arXiv:1802.03426  (2018)

\bibitem{nguyen2021wanet}
Nguyen, A., Tran, A.: Wanet--imperceptible warping-based backdoor attack. arXiv
  preprint arXiv:2102.10369  (2021)

\bibitem{nie2022diffusion}
Nie, W., Guo, B., Huang, Y., Xiao, C., Vahdat, A., Anandkumar, A.: Diffusion
  models for adversarial purification. arXiv preprint arXiv:2205.07460  (2022)

\bibitem{omarali2020virtual}
Omarali, B., Denoun, B., Althoefer, K., Jamone, L., Valle, M., Farkhatdinov,
  I.: Virtual reality based telerobotics framework with depth cameras. In: 2020
  29th IEEE International Conference on Robot and Human Interactive
  Communication (RO-MAN). pp. 1217--1222. IEEE (2020)

\bibitem{paszke2019pytorch}
Paszke, A., Gross, S., Massa, F., Lerer, A., Bradbury, J., Chanan, G., Killeen,
  T., Lin, Z., Gimelshein, N., Antiga, L., et~al.: Pytorch: An imperative
  style, high-performance deep learning library. Advances in neural information
  processing systems  \textbf{32} (2019)

\bibitem{qi2017pointnet}
Qi, C.R., Su, H., Mo, K., Guibas, L.J.: Pointnet: Deep learning on point sets
  for 3d classification and segmentation. In: Proceedings of the IEEE
  conference on computer vision and pattern recognition. pp. 652--660 (2017)

\bibitem{qi2017pointnet++}
Qi, C.R., Yi, L., Su, H., Guibas, L.J.: Pointnet++: Deep hierarchical feature
  learning on point sets in a metric space. Advances in neural information
  processing systems  \textbf{30} (2017)

\bibitem{qiu2021deepsweep}
Qiu, H., Zeng, Y., Guo, S., Zhang, T., Qiu, M., Thuraisingham, B.: Deepsweep:
  An evaluation framework for mitigating dnn backdoor attacks using data
  augmentation. In: Proceedings of the 2021 ACM Asia Conference on Computer and
  Communications Security. pp. 363--377 (2021)

\bibitem{raj2020survey}
Raj, T., Hanim~Hashim, F., Baseri~Huddin, A., Ibrahim, M.F., Hussain, A.: A
  survey on lidar scanning mechanisms. Electronics  \textbf{9}(5), ~741 (2020)

\bibitem{rajathi2023path}
Rajathi, K., Gomathi, N., Mahdal, M., Guras, R.: Path segmentation from point
  cloud data for autonomous navigation. Applied Sciences  \textbf{13}(6), ~3977
  (2023)

\bibitem{turner2019label}
Turner, A., Tsipras, D., Madry, A.: Label-consistent backdoor attacks. arXiv
  preprint arXiv:1912.02771  (2019)

\bibitem{vo2015octree}
Vo, A.V., Truong-Hong, L., Laefer, D.F., Bertolotto, M.: Octree-based region
  growing for point cloud segmentation. ISPRS Journal of Photogrammetry and
  Remote Sensing  \textbf{104},  88--100 (2015)

\bibitem{wang2019dynamic}
Wang, Y., Sun, Y., Liu, Z., Sarma, S.E., Bronstein, M.M., Solomon, J.M.:
  Dynamic graph cnn for learning on point clouds. ACM Transactions on Graphics
  (tog)  \textbf{38}(5),  1--12 (2019)

\bibitem{weber2023rab}
Weber, M., Xu, X., Karla{\v{s}}, B., Zhang, C., Li, B.: Rab: Provable
  robustness against backdoor attacks. In: 2023 IEEE Symposium on Security and
  Privacy (SP). pp. 1311--1328. IEEE (2023)

\bibitem{wen2021generative}
Wen, X., Jiang, W., Zhan, J., Bian, C., Song, Z.: Generative strategy based
  backdoor attacks to 3d point clouds: work-in-progress. In: Proceedings of the
  2021 International Conference on Embedded Software. pp. 23--24 (2021)

\bibitem{wu2021adversarial}
Wu, D., Wang, Y.: Adversarial neuron pruning purifies backdoored deep models.
  Advances in Neural Information Processing Systems  \textbf{34},  16913--16925
  (2021)

\bibitem{wu2023computation}
Wu, Y., Han, X., Qiu, H., Zhang, T.: Computation and data efficient backdoor
  attacks. In: Proceedings of the IEEE/CVF International Conference on Computer
  Vision. pp. 4805--4814 (2023)

\bibitem{wu20153d}
Wu, Z., Song, S., Khosla, A., Yu, F., Zhang, L., Tang, X., Xiao, J.: 3d
  shapenets: A deep representation for volumetric shapes. In: Proceedings of
  the IEEE conference on computer vision and pattern recognition. pp.
  1912--1920 (2015)

\bibitem{wu2020depth}
Wu, Z., Allibert, G., Stolz, C., Demonceaux, C.: Depth-adapted cnn for rgb-d
  cameras. In: Proceedings of the Asian Conference on Computer Vision (2020)

\bibitem{xiang2021backdoor}
Xiang, Z., Miller, D.J., Chen, S., Li, X., Kesidis, G.: A backdoor attack
  against 3d point cloud classifiers. In: Proceedings of the IEEE/CVF
  international conference on computer vision. pp. 7597--7607 (2021)

\bibitem{xiang2022detecting}
Xiang, Z., Miller, D.J., Chen, S., Li, X., Kesidis, G.: Detecting backdoor
  attacks against point cloud classifiers. In: ICASSP 2022-2022 IEEE
  International Conference on Acoustics, Speech and Signal Processing (ICASSP).
  pp. 3159--3163. IEEE (2022)

\bibitem{xie2023real}
Xie, X., Wei, H., Yang, Y.: Real-time lidar point-cloud moving object
  segmentation for autonomous driving. Sensors  \textbf{23}(1), ~547 (2023)

\bibitem{xu2021detecting}
Xu, X., Wang, Q., Li, H., Borisov, N., Gunter, C.A., Li, B.: Detecting ai
  trojans using meta neural analysis. In: 2021 IEEE Symposium on Security and
  Privacy (SP). pp. 103--120. IEEE (2021)

\bibitem{yoshida2020disabling}
Yoshida, K., Fujino, T.: Disabling backdoor and identifying poison data by
  using knowledge distillation in backdoor attacks on deep neural networks. In:
  Proceedings of the 13th ACM workshop on artificial intelligence and security.
  pp. 117--127 (2020)

\bibitem{zeng2021rethinking}
Zeng, Y., Park, W., Mao, Z.M., Jia, R.: Rethinking the backdoor attacks'
  triggers: A frequency perspective. In: Proceedings of the IEEE/CVF
  international conference on computer vision. pp. 16473--16481 (2021)

\bibitem{zhang2022towards}
Zhang, Y., Zhu, Y., Liu, Z., Miao, C., Hajiaghajani, F., Su, L., Qiao, C.:
  Towards backdoor attacks against lidar object detection in autonomous
  driving. In: Proceedings of the 20th ACM Conference on Embedded Networked
  Sensor Systems. pp. 533--547 (2022)

\bibitem{zheng2022global}
Zheng, Y., Li, Y., Yang, S., Lu, H.: Global-pbnet: A novel point cloud
  registration for autonomous driving. IEEE Transactions on Intelligent
  Transportation Systems  \textbf{23}(11),  22312--22319 (2022)

\end{thebibliography}
\end{document}